%% file: main.tex
\documentclass[10pt,twocolumn,letterpaper]{article}

\usepackage[pagenumbers]{cvpr} 

\usepackage{graphicx}
\usepackage{cite}
\usepackage{amsmath,amssymb,amsfonts}
\usepackage{algorithmic}
\usepackage{graphicx}
\usepackage{textcomp}
\usepackage{xcolor}
\usepackage{color}
\usepackage{colortbl}
\usepackage{xspace}
\usepackage[skip=8pt]{caption}
\usepackage[normalem]{ulem}
\newcommand{\uu}[1]{\underline{#1}}
\newcommand{\bb}[1]{\textbf{#1}}
\newcommand{\mcb}[2]{\multicolumn{#1}{c}{\textbf{#2}}}
\newcommand{\mv}{\vspace{-2mm}}
\newcommand{\point}{\item \mv}

\usepackage{bbding}
\usepackage{multirow}
\usepackage{booktabs,array}
\usepackage[utf8]{inputenc}

\newcommand{\ru}{\rule{0mm}{3mm}}

\newcommand{\NAME}{{TruFor}}
\newcommand{\NAMERES}{{Noiseprint++\xspace}}

\newcolumntype{C}[1]{>{\centering\arraybackslash}p{#1}}

\makeatletter
\newcommand\thefontsize[0]{{The current font size is: \f@size pt\par}}
\makeatother

\usepackage[pagebackref,breaklinks,colorlinks]{hyperref}

\usepackage[capitalize]{cleveref}
\crefname{section}{Sec.}{Secs.}
\Crefname{section}{Section}{Sections}
\Crefname{table}{Table}{Tables}
\crefname{table}{Tab.}{Tabs.}

\def\scalefactortab{0.90}

\begin{document}

\title{TruFor: Leveraging all-round clues for trustworthy image forgery \\ detection and localization}

\author{Fabrizio Guillaro\textsuperscript{1} \ \ \ 
Davide Cozzolino\textsuperscript{1} \ \ \ 
Avneesh Sud\textsuperscript{2} \ \ \ 
Nicholas Dufour\textsuperscript{2} \ \ \  
Luisa Verdoliva\textsuperscript{1} \\[3mm]
{\textsuperscript{1}University Federico II of Naples \ \ \ \ \ \textsuperscript{2}Google Research}}

\maketitle

\begin{abstract}
In this paper we present \NAME, a forensic framework that can be applied to a large variety of image manipulation methods, from classic cheapfakes to more recent manipulations based on deep learning. We rely on the extraction of both high-level and low-level traces through a transformer-based fusion architecture that combines the RGB image and a learned noise-sensitive fingerprint. The latter learns to embed the artifacts related to the camera internal and external processing by training only on real data in a self-supervised manner. Forgeries are detected as deviations from the expected regular pattern that characterizes each pristine image. Looking for anomalies makes the approach able to robustly detect a variety of local manipulations, ensuring generalization. In addition to a pixel-level localization map and a whole-image integrity score, our approach outputs a reliability map that highlights areas where localization predictions may be error-prone. This is particularly important in forensic applications in order to reduce false alarms and allow for a large scale analysis. Extensive experiments on several datasets show that our method is able to reliably detect and localize both cheapfakes and deepfakes manipulations outperforming state-of-the-art works. Code is publicly available at \url{https://grip-unina.github.io/TruFor/}. 
\end{abstract}

\section{Introduction}
\label{sec:intro}

\begin{figure}
    \centering
    \includegraphics[page=1, width=1.0\linewidth, trim=200 120 200 0,clip]{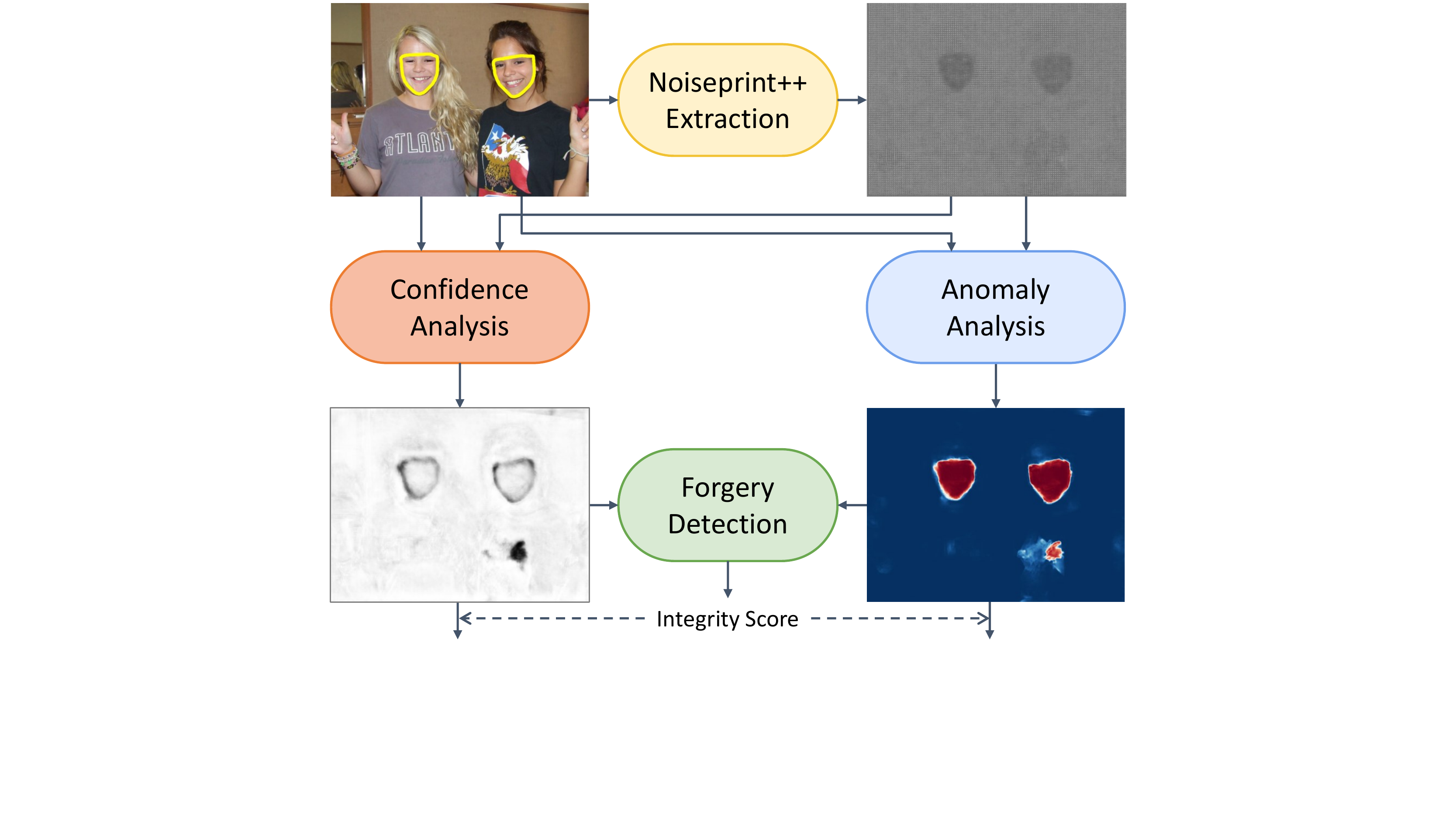}
    \caption{\NAME~ detects and localizes image forgeries (in yellow). It is based on the extraction of a learned noise-sensitive fingerprint, \NAMERES{}, which is combined with the RGB image to output an anomaly localization map. \NAMERES{} is also used jointly with the image to compute the confidence map, which estimates the less reliable regions of the anomaly heatmap (black areas), e.g. the false positive region in lower right. The confidence and anomaly maps are then used together to produce a global integrity score.}
    \label{fig:teaser}
\end{figure}

Manipulating images has never been easier, with new powerful editing tools appearing by the day. These new opportunities stimulate the creativity of benign and malicious users alike. Previously, crafting a multimedia disinformation campaign required sophisticated skills,
and attackers could do little more than copy, replicate or remove objects in an image, classic forms of image manipulations also known as ``cheapfakes''.
With the explosive growth of deep learning,
image manipulation tools have become both easier to use and more powerful,
allowing users to generate on-the-fly images of persons that do not exist or to realize credible deepfakes.
Diffusion models enable the creation of realistic image edits using natural language prompts,
photorealistically adapting the inserted manipulation to the style and lighting of the context \cite{nichol2021glide, avrahami2022blendedlatent}.

The risks posed by such tools in the wrong hands are obvious.
Indeed, in recent years there has been a growing interest on the part of governments and funding agencies in developing forensic tools capable of countering such attacks.
A major focus is on local image edits, particularly partial modifications that change the image semantics (for example the partially manipulated image in Fig.~\ref{fig:teaser}, where the two real faces have been replaced with GAN-generated ones \cite{Le2021OpenForensics}).
Multimedia forensics and related scientific fields have seen a rapid increase in activity in response to such challenges, with a large number of methods and tools proposed for image forgery detection and localization \cite{Verdoliva2020media}.
Despite considerable advances in the area, current SOTA detectors are not yet performant enough for in-the-wild deployment, due mainly to deficiencies in several areas subject to intense research:
{\it   i)} limited generalization;
{\it  ii)} limited robustness;
{\it iii)} insufficient detection performance.

Limited generalization is the inability of detectors to cope with out-of-distribution manipulations.  
Some detectors are built to exploit well-defined low-level features, e.g., traces of JPEG compression, demosaicking or interpolation \cite{bianchi2011improved, Park2018double, Bammey2020adaptive}, while others are typically developed to work well only on specific types of manipulations, like splicing \cite{Salloum2018image, Kwon2020catnet}.
In addition, in a realistic scenario  
images also undergo numerous forms of non-malicious degradation,
(e.g. recompression, resizing, etc) - also called \emph{laundering}.
For example, social networks compress and resize uploaded images, both of which can easily remove forensic traces.
Finally, most SOTA methods perform image forgery localization, leaving detection as an afterthought \cite{cozzolino2022datadriven}, which is typically derived as a global integrity score from the localization heatmap itself \cite{Huh2018, Wu2019mantra, Rao2021multi}. Few methods address the detection task directly
\cite{Marra2020, Zhang2020dense, wang2022objectformer, chen2021image}.
As a result, detection accuracy is poor, with a high false alarm rate.
In a realistic setting where manipulated images are rare, such performance could cause more problems than it solves, with false positives drastically outnumbering true positives. 

This work addresses such shortcomings, with a focus on robust detection under varied manipulations.
Our aim is to first establish whether the image under analysis has been manipulated or not, and subsequently consider forgery localization only for images where a forgery has been detected. To perform in a real-world scenario where images undergo many post-processing steps that may attenuate forensic traces, our design was guided by the need to leverage information at multiple scales (both low and high-level features) even in complex scenarios. Our framework estimates a confidence map that associates localization results with region-specific uncertainty, allowing many potential false alarms to be rejected. The block diagram of our method is presented in Fig.~\ref{fig:teaser}. Overall, in this work we make the following key contributions:
\begin{itemize}
\point we propose a new framework, \NAME, which outputs a global integrity score, an anomaly-based localization map and an associated confidence map;
\point we propose a new noise-sensitive fingerprint, \NAMERES{}, with enhanced robustness to image laundering;
\point we combine low-level and high-level evidence to perform anomaly analysis, which together with the confidence analysis provide more reliable decisions;
\point we carry out extensive experiments on several benchmarks, considering new and challenging scenarios, and demonstrate that our method achieves state-of-the-art performance in both detection and localization tasks.
\end{itemize}

\section{Related Work} 
\label{sec:related}

\noindent
\bb{Forensic artifacts.}
Low-level artifacts are caused by the in-camera acquisition process, such as the sensor, the lens, the color filter array or the JPEG quantization tables. In all cases, these are very weak traces, that can be highlighted by suppressing the image content by means of high-pass filters or denoising. The most common filters used for this task are the spatial rich models (SRM) \cite{Fridrich2012rich}, often included as a pre-processing step in some CNN models for forensic analysis. In \cite{Rao2016} a set of around 30 fixed high-pass filters are used, instead in \cite{Bayar2016} the high-pass filters are learnt during training. These fixed and trainable filters have been used in many other subsequent works to perform a noise sensitive analysis \cite{zhou2018learning, Wu2019mantra, Hu2020span, chen2021image, Yang2020constrained}. 
A different perspective is considered in \cite{Cozzolino2020noise}, where the extraction of low-level artifacts is carried out by learning a sort of ``camera model fingerprint'', the noiseprint, that bears traces of in-camera processing steps. When a manipulation is present, the noiseprint structure is absent and this anomaly is interpreted as a forgery. In this work we leverage noiseprint and further enhance it so as to make it work in more challenging scenarios. 

\begin{figure*}[t!]
    \centering
    \includegraphics[width=1.0\linewidth, trim=110 295 110 0, clip, page=2]{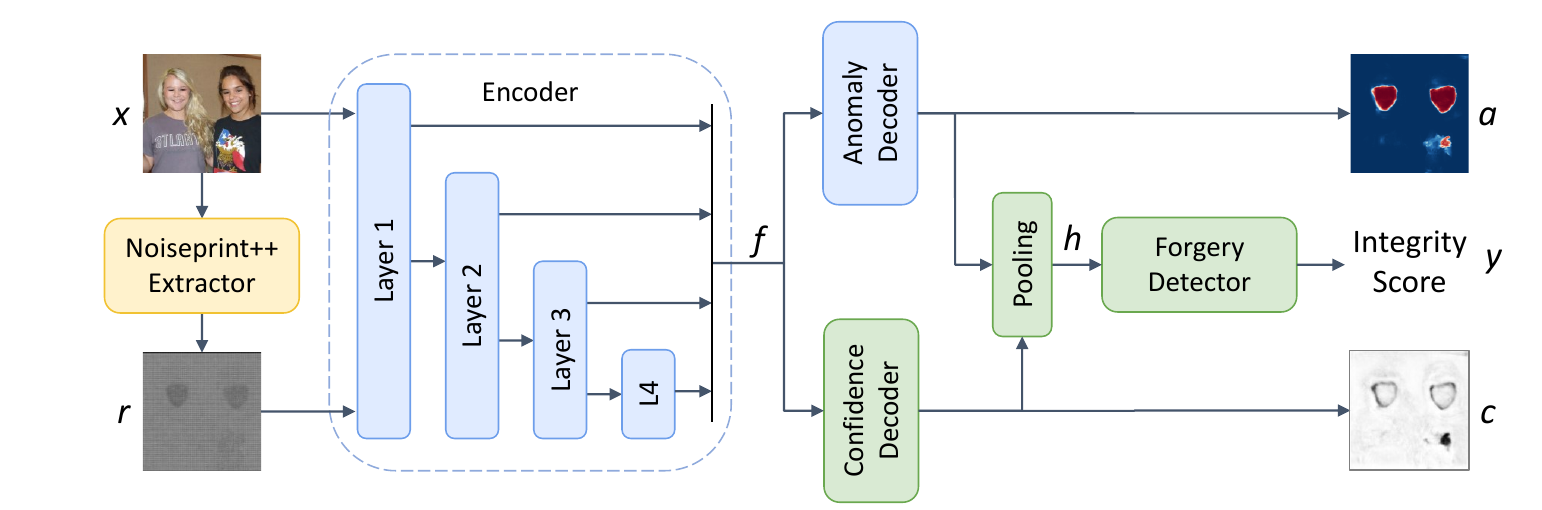}
    \caption{TruFor framework. The \NAMERES~extractor takes the RGB image to obtain a learned noise-sensitive fingerprint. The encoder uses both the RGB input and \NAMERES{} for jointly computing the features that will be used by the anomaly decoder and the confidence decoder for pixel-level forgery localization and confidence estimation, respectively. The forgery detector exploits the localization map and the confidence map to make the image-level decision. The different colors identify the modules learned in each of the three training phases.}
    \label{fig:scheme}
\end{figure*}

In general, low-level features are combined with high-level ones to carry out a more effective detection. Pioneering work in the field is the two-branch approach proposed in \cite{zhou2018learning}, where the features of the noise and RGB stream are combined together through bilinear pooling. 
Other works also propose late fusion \cite{chen2021image}, 
while others \cite{Wu2019mantra, Hu2020span, wang2022objectformer} perform early fusion or even middle fusion \cite{kwon2022learning}.
We belong to this last category, but use an approach that fuses noise and RGB channels using cross-modal feature calibration \cite{liu2022cmx}.

\vspace{2mm}
\noindent
\bb{Forgery detection vs localization.}
The majority of the state-of-the-art methods focus on image localization, with architectures often inspired by semantic segmentation, and detection is a byproduct of such analysis \cite{cozzolino2022datadriven}. The integrity score is computed by a suitable post-processing of the localization heatmap aimed at extracting a global decision statistic, such as the average or the maximum value of the heatmap \cite{Huh2018, Wu2019mantra, Bi2019rru}.
Only a few works explicitly treat the detection problem.
In particular, some recent approaches \cite{Zhang2020dense, chen2021image, wang2022objectformer, liu2022pscc} jointly train the model both for localization and detection 
through suitable losses at image-level. 
In \cite{Zhang2020dense, wang2022objectformer}
global average pooling is applied to the middle features, while in \cite{chen2021image}
max average pooling is carried out on the localization heatmap.
A different perspective can be found in \cite{Marra2020}, where it is proposed to analyze the whole image avoiding resizing (so as not to lose precious forensics traces) through a gradient checkpointing technique, that helps for the joint optimization of patch-level feature extraction and image-level decision. 

Different from current literature, in this paper we explicitly design a forgery detection module that takes as input the anomaly-based map and the confidence map. This additional input is crucial to reduce the number of false alarms on pristine data and provide a more trustworthy tool.

\vspace{2mm}
\noindent
\bb{Reliability in multimedia forensics}
Designing reliable detectors is important in several computer vision applications, 
however, it is even more critical for our task, since forensic traces are often imperceptible to visual inspection. 
The problem is even more relevant when deep learning based methods are used, since image forensics tools are challenged by out-of-distribution data \cite{Verdoliva2020media}.
In the context of JPEG artifacts and resampling analysis, initial efforts to develop reliable forensics detectors are carried out in \cite{Lorch2020reliable, Mayer2020toward}, where it is proposed to use Bayesian neural networks that provide an uncertainty range with every prediction. In this way, the user can quantify trust on the final prediction. 

Inspired by \cite{corbiere2019addressing}, our work aims at making a further step in this direction and proposes a method using external uncertainty quantification \cite{gawlikowski2021survey} to design a confidence map from the anomaly localization heatmap.

\section{Method}
\label{sec:method}

In this Section we begin by presenting an overview of \NAME, which is illustrated in Fig.~\ref{fig:scheme}. Subsequent subsections will provide the details of each component.
First of all, from the input RGB image, $x$, we extract its \NAMERES{}, $r=\mathcal{R}(x)$, a learned noise-sensitive fingerprint of the same resolution as $x$.
Then, both $x$ and $r$ feed two networks that extract the anomaly map $a$ and the confidence map $c$ of the image.
These networks have the same encoder-decoder architecture, with a shared encoder that extracts suitable dense features, $f=\mathcal{E}(x,r)$, which are processed
by the anomaly decoder to extract the anomaly map, $a=\mathcal{D_A}(f)$, and
by the confidence decoder to extract the confidence map, $c=\mathcal{D_C}(f)$.
The information gathered in the anomaly map is summarized in a compact descriptor, $h=\mathcal{P}(a,c)$,
by means of a weighted pooling block, with weights depending on the confidence information.
Finally, this descriptor is processed by a classifier which computes an integrity score, $y=\mathcal{C}(h)$.

Integrity score, anomaly map and confidence map are all provided to the final user for further analyses.
At a first level,
only the integrity score is necessary to perform automated forgery detection.
In case a fake is detected, 
the user can dive deeper using the anomaly map to identify  manipulated suspected regions, along with the confidence map to distinguish valid predictions of forged regions from random anomalies.
For pristine images, instead,
the anomaly map does not localize possible forgeries but only random statistical anomalies, and should be discarded.

\subsection{\NAMERES}

\noindent
\bb{Motivation.}
Digital images are marked by a long trail of subtle, invisible traces.
These may have many distinct origins,
from the unavoidable imperfections of the camera hardware,
to the in-camera processing steps of image acquisition, 
to all the out-camera processes encountered by the image during its lifetime.
When images are manipulated, these telltale traces may be corrupted,
an event that, if detected, allows one to carry out powerful forensic analyses.

\begin{figure}[t!]
    \centering
    \includegraphics[width=1.0\linewidth, trim=110 185 110 0, clip, page=3]{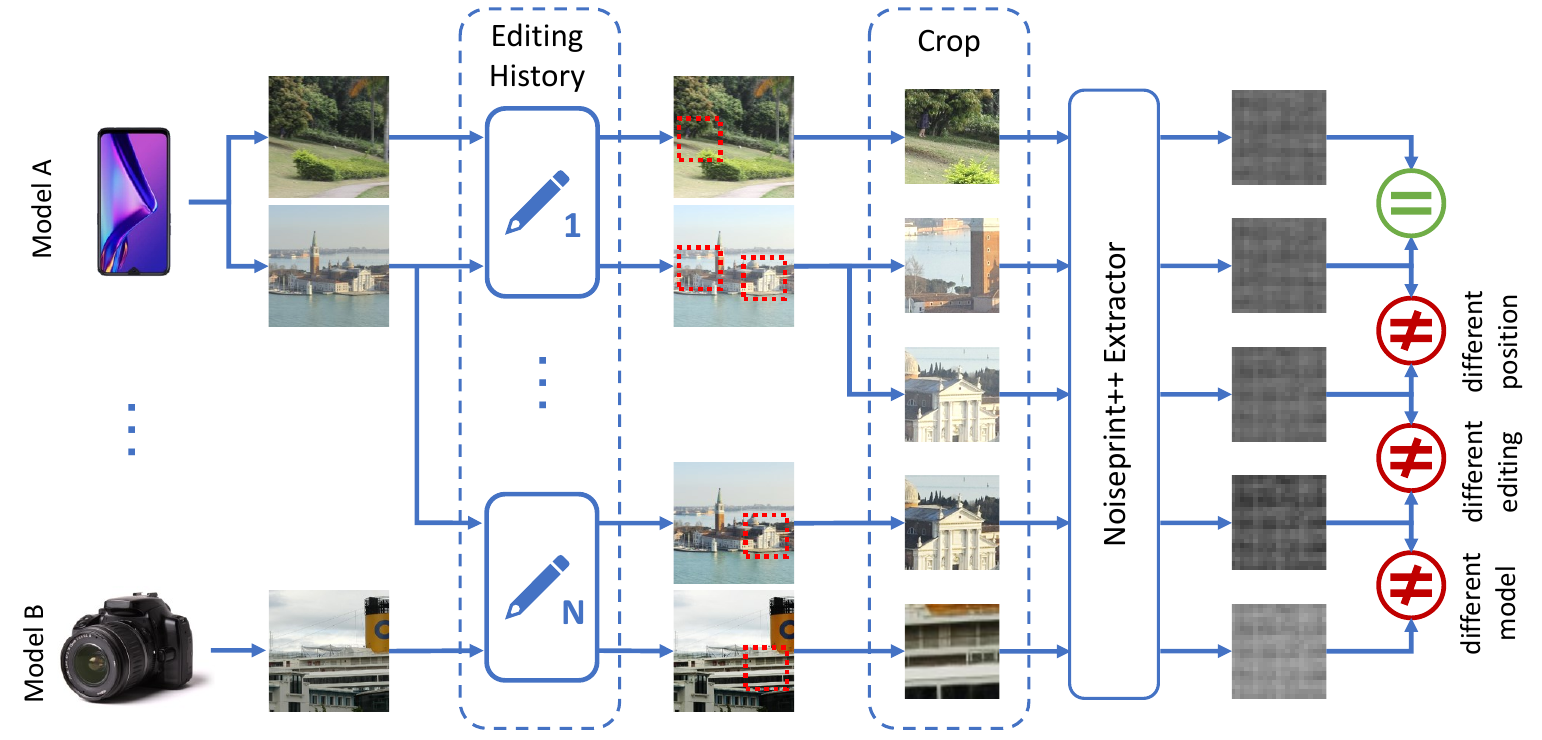}
    \caption{\NAMERES{} training procedure. Each image is subjected to a different combination of processing operations, namely \textit{editing history}. Different crops are extracted from real images taken from many different cameras. During training, the distance between the outputs is minimized for patches coming from the same camera model, same position and same editing history.}
    \label{fig:training_residual}
\end{figure}

In \cite{Cozzolino2020noise}
a deep learning-based method has been proposed to extract from each image its noiseprint,
an image-size pattern where all traces related to in-camera processing steps are collected and emphasized.
This is trained in a self-supervised manner using only pristine images.
While this ensures it can be trained on a large corpus,  
it shows limited robustness to image impairments induced by out-camera processes. This is a significant shortcoming,
considering that many forms of impairments are possible during the lifetime of an image.
To overcome this limitation, we propose \NAMERES{}, an improved image fingerprint
which highlights traces related not only to in-camera but also to out-camera processes.
In other words, \NAMERES{} captures information not only on the camera model but also on its editing history,
improving its reliability.

\vspace{2mm}
\noindent
\bb{Self-supervised contrastive learning.}
The proposed \NAMERES{} extractor learns patch-level self-similarities by means of contrastive learning.
Similar to \cite{Cozzolino2020noise},
we adopt the DnCNN architecture \cite{Zhang2017beyond} with $15$ trainable layers, $3$ input channels, $1$ output channel.
The extractor is trained on patches of $64\times 64$ pixels randomly extracted from images of the dataset.
Training is aimed at obtaining
the same noise-sensitive fingerprint for patches that share the same properties
and different noise residuals for patches that are different under some respect.
Figure~\ref{fig:training_residual}, in particular,
highlights that two patches are considered different, and hence characterized by different noise residuals, when they
{\it   (i)} come from different sources;
{\it  (ii)} are drawn from different spatial positions;
{\it (iii)} have different editing histories.
These constraints, in turn, aim at telling apart patches
{\it   (i)} generated by different cameras,
{\it  (ii)} moved from one spatial location to another and
{\it (iii)} coming from images that have been differently post-processed.
This latter property, in particular, distinguishes \NAMERES{} from its ancestor and improves its effectiveness.
We adopt the InfoNCE contrastive loss~\cite{khosla2020supervised}:
\begin{equation}
    \mathcal{L}_{contr} =  - \sum_{i \in \mathcal{B}} \log
    \frac{\sum_{j \in \mathcal{N}_i}  e^{-s(i,j)}}
         {\sum_{j \in  \mathcal{B}-\{i\} } e^{-s(i,j)} }
    \label{equ:contrastive}
\end{equation}
where
$\mathcal{B}$ is a batch of patches,
$s(i,j)$ is the squared Euclidean distance between $i$-th and $j$-th residual patches, and
$\mathcal{N}_i$ is the subset of patches with the same origin, position and editing history as the $i$-th patch.
During contrastive learning, we introduce a large variety of possible editing operations,
such as resizing, compression and illumination changes, for a total of 512 different history pipelines.

\begin{figure}[t!]
    \centering
    \includegraphics[page=1, width=1.0\linewidth, trim=170 210 170 0,clip]{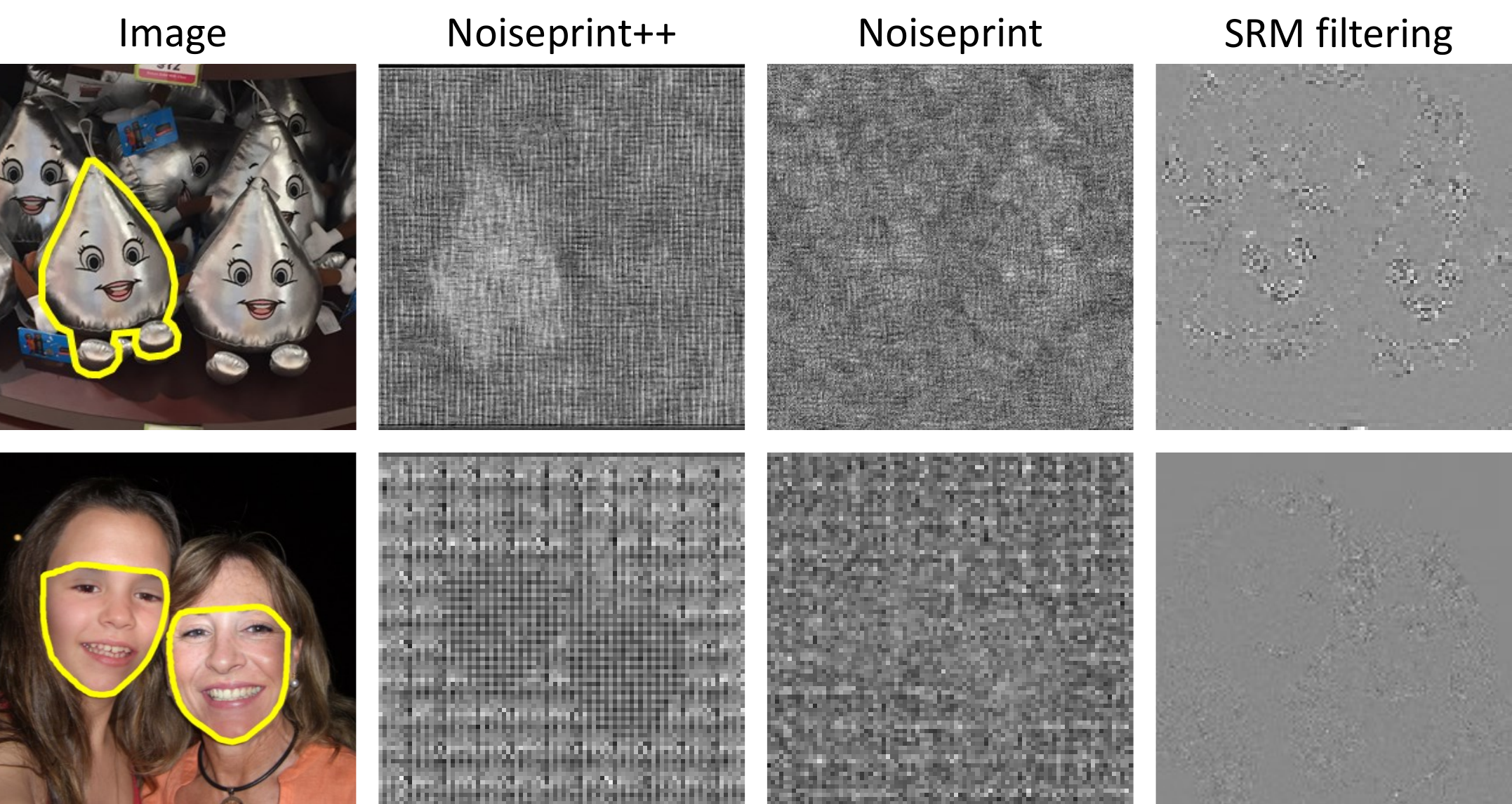}
    \caption{From left to right: manipulated image, \NAMERES{}, Noiseprint and residual obtained through SRM-based filtering. We can notice that forensic artifacts are much more enhanced using our learned noise-sensitive fingerprint. In particular, we can observe the typical $8\times 8$ grid that characterizes JPEG compressed images
    and \NAMERES{} can highlight the grid inconsistencies over the forged area better than Noiseprint.
}
    \label{fig:residuals}
\end{figure}

\begin{figure}[t!]
    \centering
    \includegraphics[page=2, width=1.0\linewidth, trim=0 310 0 0,clip]{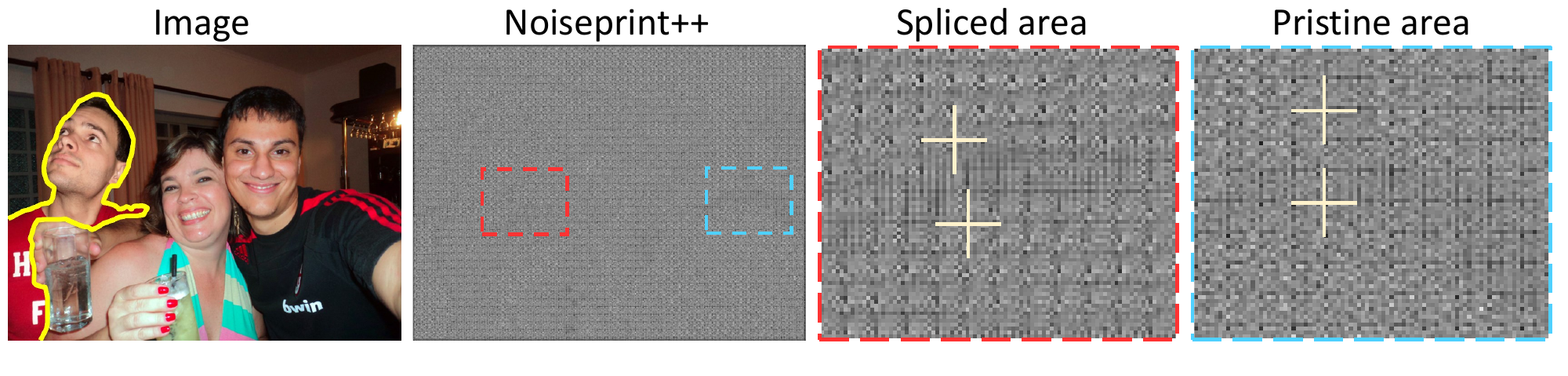}
    \caption{Example of a manipulated image and its \NAMERES{}, which clearly shows a JPEG grid misalignment caused by the composition. This is visible in the spliced area (red), but not in the pristine one (blue), as highlighted in the zoomed regions where $+$ indicates a JPEG grid boundary.}
    \label{fig:zoom2}
\end{figure}

In Fig.~\ref{fig:residuals} we show two examples of \NAMERES{}
compared to noiseprint and some standard spatial domain residuals (SRM filters), while in Fig.~\ref{fig:zoom2} we show a manipulated image where we can notice a JPEG grid misalignment in correspondence to the forged area.

\subsection{Anomaly localization map}
We treat the forgery localization task as a supervised binary segmentation problem and combine the \NAMERES{} information with the high-level features from the RGB image.
To this end, we adopt the CMX architecture \cite{liu2022cmx},
a cross-modal fusion framework originally designed for multi-modal semantic segmentation, but easily generalizable to other tasks.
Features from input image and \NAMERES{}
are extracted on two parallel branches which have a shared encoder architecture from a semantic segmentation method.
In particular, we rely on SegFormer \cite{xie2021segformer}, a hierarchical network based on a Transformer encoder.
Interaction is carried out between each stage using a Cross-Modal Feature Rectification Module,
which calibrates the information coming from one modality using features extracted from the other modality.
The calibration helps to filter out noisy information of a modality using the knowledge of the other modality.
The rectified features of both modalities are provided as input to the Feature Fusion Module,
which uses a cross-attention mechanism to merge them into a single feature map.
The fused feature maps of all stages represent the input of the decoder, which is used to generate the final anomaly map.
For the decoder, we keep the lightweight multilayer perceptron used in SegFormer \cite{xie2021segformer}. Details are provided in the Supplementary.

\newcommand{\minus}{\!-\!}
During phase 2 training, the loss function is a combination of the weighted cross-entropy and the dice loss \cite{Milletari2016vnet}:
\begin{equation}
    \mathcal{L}_2 = \lambda_{ce} \mathcal{L}_{ce} + (1-\lambda_{ce}) \mathcal{L}_{dice}
\end{equation}
with $\lambda_{ce}$ set experimentally to 0.3.
The  weighted cross-entropy loss is defined as
\begin{equation}
    \mathcal{L}_{ce} = -\frac{1}{N} \sum_i \gamma_0 (1 \minus g_i) \log(1 \minus a_i) + \gamma_1 g_i \log a_i
\end{equation}
with $g_i$ and $a_i$ the $i$-th pixel of the ground truth and estimated anomaly maps respectively,
and $N$ the number of pixels in the image.
The weights $\gamma_0$ and $\gamma_1$, are set to $0.5$ and $2.5$
to take into account the imbalance between pristine and fake pixels in the training-set.

\setlength{\tabcolsep}{2pt}
\begin{table*}[t]
	\centering
	\small
	\scalebox{\scalefactortab}{
	\begin{tabular}{rC{0.8cm}C{0.8cm}C{0.8cm}C{0.8cm}C{0.8cm}C{0.8cm}C{0.8cm}C{0.8cm}C{0.8cm}C{0.8cm}C{0.8cm}C{0.8cm}C{0.8cm}C{0.8cm}C{0.8cm}C{0.8cm}C{0.8cm}C{0.8cm}} 
		\toprule
		\ru \multirow{2}{*}{\bb{Method}}              &\mcb{2}{CASIAv1+}      &\mcb{2}{Coverage}      &\mcb{2}{Columbia}      &\mcb{2}{NIST16}
		                                             &\mcb{2}{DSO-1}         &\mcb{2}{VIPP}          &\mcb{2}{OpenFor.}      &\mcb{2}{CocoGlide}     & \mcb{2}{AVG}          \\ 
		\ru                                           & best      & fixed     & best      & fixed     & best      & fixed     & best      & fixed 
		                                             & best      & fixed     & best      & fixed     & best      & fixed     & best      & fixed     & best    & fixed       \\
		\cmidrule(lr){1-1} \cmidrule(lr){2-3} \cmidrule(lr){4-5} \cmidrule(lr){6-7} \cmidrule(lr){8-9} \cmidrule(lr){10-11} \cmidrule(lr){12-13} \cmidrule(lr){14-15} \cmidrule(lr){16-17} \cmidrule(lr){18-19}
		\ru  ADQ         \cite{bianchi2011improved}   & .494      & .302      & .167      & .165      & .401      & .401      & .238      & .146     
		                                             & .483      & .421      & .549      & .457      & .644      & .414      & .302      & .300      & .410     & .326       \\
		\ru  Splicebuster \cite{Cozzolino2015splice}  & .252      & .143    & .321      & .192         & .811      & .565         & .312      & .174        
		                                             & .662      & .372       & .432      & .260         & .459      & .340         & .434      & .332         & .460      & .297         \\ 
	    \ru  EXIF-SC      \cite{Huh2018}              & .255      & .106      & .332      & .164      & .880      & .798      & .298      & .227    
	                                                 & .577      & .442      & .424      & .215      & .318      & .175      & .424      & .293      & .437      & .303      \\
		\ru  CR-CNN       \cite{Yang2020constrained}  & .538      & .481      & .487      & .391      & .779      & .631      & .363      & .300     
	                                                 & .377      & .289      & .355      & .282      & .143      & .110      & .577      & .447      & .452      & .366      \\
		\ru  RRU-Net      \cite{Bi2019rru}            & .498      & .408      & .339      & .279      & .629      & .575      & .218      & .154     
	                                                 & .360      & .312      & .336      & .272      & .206      & .157      & .504      & .416      & .386      & .322      \\
		\ru  ManTraNet    \cite{Wu2019mantra}         & .320      & .180      & .486      & .317      & .650      & .508      & .225      & .172     
	                                                 & .537      & .412      & .373      & .255      & .661      & .551      & .673      & \uu{.516} & .491      & .364      \\
		\ru  SPAN         \cite{Hu2020span}           & .169      & .112      & .428      & .235      & .873      & .759      & .363      & .228     
	                                                 & .390      & .233      & .375      & .223      & .176      & .089      & .350      & .298      & .391      & .272      \\
		\ru  AdaCFA       \cite{Bammey2020adaptive}   & .158      & .128      & .215      & .183      & .587      & .403      & .124      & .106    
	                                                 & .262      & .235      & .210      & .184      & .115      & .098      & .357      & .314      & .254      & .206      \\
		\ru  CAT-Net v2   \cite{kwon2022learning}     & \bb{.852} & \bb{.752} & .582      & .381      & \bb{.923} & \bb{.859} & .417      & .308     
	                                                 & .673      & \uu{.584} & \uu{.672} & \uu{.590} & \bb{.947} & \bb{.899} & .603      & .434      & \uu{.709} & \uu{.601} \\
		\ru  IF-OSN       \cite{wu2022robust}         & .676      & .553      & .472      & .304      & .836      & .753      & \uu{.449} & \uu{.330}    
	                                                 & .621      & .470      & .508      & .403      & .204      & .123      & .589      & .428      & .544      & .421      \\
		\ru  MVSS-Net     \cite{chen2021image}        & .650      & .528      & \uu{.659} & \uu{.514} & .781      & .729      & .372      & .320      
	                                                 & .459      & .358      & .485      & .389      & .225      & .117      & .642      & .486      & .534      & .430      \\
		\ru  PSCC-Net     \cite{liu2022pscc}          & .670      & .520      & .615      & .473      & .760      & .604      & .210      & .113      
	                                                 & .733      & .458      & .309      & .183      & .353      & .105      & \uu{.685} & .515      & .542      & .371      \\
		\ru  Noiseprint   \cite{Cozzolino2020noise}   & .205      & .137      & .342      & .229         & .835      & .513         & .345      & .196        
		                                             & \uu{.811} & .439         & .546      & .382         & .675      & .420         & .405      & .318         & .521      & .329        \\
		\ru  \NAME~(ours)                             & \uu{.822} & \uu{.737} & \bb{.735} & \bb{.600} & \uu{.914} & \bb{.859} & \bb{.470} & \bb{.399}
		                                             & \bb{.973} & \bb{.930} & \bb{.746} & \bb{.693} & \uu{.901} & \uu{.827} & \bb{.720} & \bb{.523} & \bb{.785} & \bb{.696} \\
		\bottomrule       
	\end{tabular}    
	}
	\caption{Pixel-level F1 performance of image forgery localization. Results are shown for the metric computed using the best threshold per image and using a fixed threshold (0.5). First and second rankings are shown in bold and underlined respectively. For the fixed threshold, Splicebuster and Noiseprint have been evaluated after a Normalization between 0 and 1, since they provide maps in arbitrary ranges.}
	\label{tab:comparisonF1}
\end{table*}

\subsection{Confidence map and integrity score}
Many SoTA methods perform localization first, and then use some global statistics of the localization map to perform detection.
We also need global
statistics about anomalies, 
but the anomaly map cannot be blindly trusted, as it highlights both manipulated areas and pristine areas with unusual statistics. Hence we propose a method to compute a per-pixel confidence estimate of the predicted anomaly map, which is used to compute robust global statistics for detection.
In the pooling block we compute four {\em weighted} statistics of the anomaly map, maximum, minimum, average, and mean square,
where the weights are drawn from the confidence map and help de-emphasize pristine anomalous areas of the image.
In formulas
\begin{eqnarray}
    a_{\rm avg} = \sum_i \acute{c}_i \, a_i   ; & \hspace{-2mm} a_{\rm max} =  \log{ \sum_i \acute{c}_i \, e^{ a_i} }  \\
    a_{\rm msq} = \sum_i \acute{c}_i \, a_i^2 ; &\,\,\,\, a_{\rm min} = -\log{ \sum_i \acute{c}_i \, e^{-a_i} } 
\end{eqnarray}
where $a_i$ and $\acute{c}_i$ are the values of the anomaly and confidence maps at pixel $i$, respectively, 
the latter normalized to unit sum,
and we adopt a smooth approximation of the minimum and maximum functions.
To these features we add the four corresponding features extracted from the confidence map $c_{\rm avg}, c_{\rm msq}, c_{\rm max}, c_{\rm min}$ 
obtaining eventually a 8-component feature vector, $h$, which is used to predict the integrity score $y$. 

The confidence and anomaly maps are generated in parallel, by decoding the same input features with two decoders having the same architecture, as done in \cite{corbiere2019addressing}.
However, while the anomaly values point out statistical outliers,
confidence values have to recognize which anomaly values can be trusted.
Hence, the confidence decoder must be trained with suitable {\it ad hoc} reference data.
To this end, we use another map, $t$, the true class probability map \cite{corbiere2019addressing}:
\begin{equation}
    t_i =  (1-g_i) \, (1-a_i) + g_i \, a_i
\end{equation}
where $g_i$ and $a_i$ are the pixel values of the localization ground truth and of the anomaly map.
The ground truth values, $g_i$, are 1 for manipulated pixels and 0 for pristine ones.
Therefore, the true class probability map is close to 1
when large anomaly values occur for manipulated pixels or small anomaly values occur for pristine pixels.
Instead, it is close to 0 when manipulated pixels are not seen as anomalous or anomalies are detected in pristine data.
This latter case is especially important as it may easily lead to false alarms.
The confidence decoder must learn to identify and discard these wrong pieces of information.
Hence, the confidence loss, $\mathcal{L}_{conf}$, is defined as the mean squared error between the predicted confidence map $c$ and its reference $t$.

Finally, to maximize the system reliability, the confidence decoder is trained jointly with the final binary classifier.
Therefore we train this phase using a weighted sum of confidence loss and detection loss
\begin{equation}
    \mathcal{L}_3 = \mathcal{L}_{conf} + \lambda_{det} \mathcal{L}_{det}
\end{equation}
where $\mathcal{L}_{det}$ is the balanced cross-entropy on the predicted image-level integrity score $y$ and $\lambda_{det}$ is set to 0.5.

\section{Results}
\label{sec:results}

\subsection{Experimental Setup}

\noindent
\bb{Training.}
Our approach includes three separate training steps.
First, we train the \NAMERES\ extractor using a large dataset of pristine images 
publicly available on two popular photo-sharing websites: Flickr (www.flickr.com) and DPReview (www.dpreview.com). 
The whole dataset contains 24,757 images acquired from 1,475 different camera models (8 to 92 images per model) of 43 brands. 
Then, we train encoder and decoder of the anomaly localization network using the same datasets as proposed in CAT-Net v2 \cite{kwon2022learning}, 
comprising pristine and fake images with the corresponding ground truths.
Finally, using this same dataset, 
we train the confidence map decoder and the forgery detector.
More details on these datasets can be found in the supplementary.

\vspace{2mm}
\noindent
\bb{Testing.}
We benchmarked our model on seven publicly available datasets and one more dataset of local manipulations created by us using diffusion models.
More specifically, we use 
CASIA v1 \cite{Dong2013}, 
Coverage \cite{wen2016coverage},
Columbia \cite{hsu2006detecting}, 
NIST16 \cite{Guan2019MFC},
DSO-1 \cite{Carvalho2013}, and 
VIPP \cite{bianchi2012image}, 
which are extensively used in the literature and include cheapfakes manipulations, like splicing, copy-move and inpainting. 
Overall these datasets comprise a total of 1530 fake images and 1412 real ones.
Then, we added
OpenForensics \cite{Le2021OpenForensics}
a large dataset of face manipulations generated using GAN models, from which we sampled 2000 images, and
CocoGlide, including 512 images we generated from the COCO 2017 validation set \cite{lin2014microsoft} using the GLIDE diffusion model \cite{nichol2021glide}.

\setlength{\tabcolsep}{2pt}
\begin{table*}
	\centering
	\small
	\scalebox{\scalefactortab}{
	\begin{tabular}{rC{0.8cm}C{0.8cm}C{0.8cm}C{0.8cm}C{0.8cm}C{0.8cm}C{0.8cm}C{0.8cm}C{0.8cm}C{0.8cm}C{0.8cm}C{0.8cm}C{0.8cm}C{0.8cm}C{0.8cm}C{0.8cm}} 
		\toprule		    
		\ru \multirow{2}{*}{\bb{Method}}              &\mcb{2}{CASIAv1+}      &\mcb{2}{Coverage}      &\mcb{2}{Columbia}      &\mcb{2}{NIST16}
		                                             &\mcb{2}{DSO-1}         &\mcb{2}{VIPP}          &\mcb{2}{CocoGlide}    & \mcb{2}{AVG}       \\ 
		\ru                                           & AUC       & Acc     & AUC      & Acc     & AUC      & Acc     & AUC    & Acc 
		                                             & AUC       & Acc     & AUC      & Acc     & AUC      & Acc     & AUC    & Acc   \\
		\cmidrule(lr){1-1} \cmidrule(lr){2-3} \cmidrule(lr){4-5} \cmidrule(lr){6-7} \cmidrule(lr){8-9} \cmidrule(lr){10-11} \cmidrule(lr){12-13} \cmidrule(lr){14-15} \cmidrule(lr){16-17}
		\ru  ADQ         \cite{bianchi2011improved}   & .816       & .523       & .495      & .495      & .500      & .500      & .484      & .503 
		                                             & .569       & .560       & .736      & .551      & .496      & .496      & .585      & .518   \\
		\ru  Splicebuster \cite{Cozzolino2015splice}  & .406       &  -         & .541      &  -        & .597      &  -        & .610      &  -  
		                                             & .751       &  -         & .539      &  -        & .529      &  -        & .568      &  -     \\ 
	    \ru  EXIF-SC      \cite{Huh2018}              & .490       & .500       & .498      & .500      & .976      & .506      & .504      & .500     
		                                             & .764       & .500       & .617      & .500      & .526      & .500      & .625      & .501   \\
		\ru  CR-CNN       \cite{Yang2020constrained}  & .670       & .535       & .553      & .510      & .755      & .628      & .737      & \uu{.641}   
		                                             & .576       & .535       & .504      & .558      & .589      & .533      & .626      & .563   \\
		\ru  RRU-Net      \cite{Bi2019rru}            & .574       & .488       & .482      & .500      & .583      & .500      & .666      & .500    
		                                             & .444       & .500       & .534      & .500      & .533      & .503      & .545      & .499   \\
		\ru  ManTraNet    \cite{Wu2019mantra}         & .644       & .500       & \uu{.760} & .500      & .810      & .500      & .624      & .500    
		                                             & \uu{.874}  & .500       & .530      & .500      & \bb{.778} & .500      & .717      & .500   \\
		\ru  SPAN         \cite{Hu2020span}           & .480       & .487       & .670      & .605      & \bb{.999} & \uu{.951} & .632      & .597     
		                                             & .669       & .510       & .580      & \uu{.572} & .475      & .491      & .644      & .602   \\
		\ru  AdaCFA       \cite{Bammey2020adaptive}   & .500       & .500       & .500      & .500      & .500      & .500      & .500      & .500     
		                                             & .500       & .500       & .500      & .500      & .500      & .500      & .500      & .500   \\
		\ru  CAT-Net v2   \cite{kwon2022learning}     & \bb{.942 } & \bb{.838 } & .680      & \uu{.635} & .977      & .803      & \uu{.750} & .597     
		                                             & .747       & .525       & \uu{.813} & .565      & .667      & .580 & \uu{.797} & \uu{.649}   \\
		\ru  IF-OSN       \cite{wu2022robust}         & .735       & .635       & .557      & .510      & .882      & .522      & .658      & .553      
		                                             & .853       & .505       & .696      & .522      & .611      & .567      & .713      & .545   \\
		\ru  MVSS-Net     \cite{chen2021image}        & \uu{.932}  & .808       & .733      & .545      & .984      & .667      & .579      & .538      
		                                             & .552       & .485       & .629      & .522      & .654      & .536      & .723      & .586   \\
		\ru  PSCC-Net     \cite{liu2022pscc}          & .869       & .683       & .657      & .550      & .300      & .508      & .485      & .456      
		                                             & .650       & .543       & .574      & .507      & \uu{.777} & \bb{.661} & .616      & .558   \\
		\ru  E2E          \cite{Marra2020}            & .377       & .433       & .494      & .505      & .894      & .639      & .718      & .603      
		                                             & .803       & \uu{.565}  & .617      & .543      & .530      & .525      & .633      & .545   \\
		\ru  Noiseprint   \cite{Cozzolino2020noise}   & .494       &  -         & .525      &  -        & .872      &  -        & .618      &  -  
		                                             & .821       &  -         & .580      &  -        & .520      &  -        & .633      &  -     \\ 
		\ru  \NAME~(ours)                             & .916       & \uu{.813}  & \bb{.770} & \bb{.680} & \uu{.996} & \bb{.984} & \bb{.760} & \bb{.662} 
		                                             & \bb{.984}  & \bb{.930}  & \bb{.820} & \bb{.761} & .752 & \uu{.639} & \bb{.857} & \bb{.781}   \\ 
		\bottomrule    
	\end{tabular}
	}
	\caption{Image-level AUC and balanced Accuracy performance of image forgery detection. Splicebuster and Noiseprint cannot be evaluated using a fixed threshold because they provide maps in arbitrary ranges.}

	\label{tab:comparisonAUC}
\end{table*}

\vspace{2mm}
\noindent
\bb{Metrics.}
As in most of the previous works, 
we measure pixel-level performance in terms of F1, and report results using both the best threshold and the default 0.5 threshold. 
Instead, for image-level analysis we use AUC, which does not require setting a decision threshold, 
and balanced accuracy, which takes into account both false alarms and missed detection, in which case the threshold is set again to 0.5.

\newcommand{\mc}[1]{\multicolumn{6}{l}{\rule{0mm}{5mm} {\bf #1}}\\ \hline}
\setlength{\tabcolsep}{3pt}
\begin{table}[b]
\centering
{\footnotesize
\begin{tabular}{rccccc}
            \toprule
            \ru                                          &              & \mcb{2}{Input Type}               & \mcb{2}{Task}           \\
            \cmidrule(lr){3-4} \cmidrule(lr){5-6}
            \ru    Acronym [ref]                         & Artifact     & RGB        & Other                & L          & D          \\ \midrule
            \ru ADQ           \cite{bianchi2011improved} & JPEG         & \checkmark & DCT analysis         & \checkmark & ~          \\
            \ru Splicebuster  \cite{Cozzolino2015splice} & camera-based &            & fixed HP filter      & \checkmark & ~          \\
            \ru EXIF-SC       \cite{Huh2018}             & camera-based & \checkmark &   -                  & \checkmark & \checkmark \\
            \ru AdaCFA        \cite{Bammey2020adaptive}  & demosaicing  & \checkmark &    -                 & \checkmark &            \\ 
            \ru Noiseprint    \cite{Cozzolino2020noise}  & camera-based & \checkmark &   -                  & \checkmark & ~          \\
            \ru ManTraNet     \cite{Wu2019mantra}        & editing      & \checkmark & HP filters           & \checkmark & \checkmark \\
            \ru RRU-Net       \cite{Bi2019rru}           & splicing     & \checkmark &  -                   & \checkmark & \checkmark \\ 
            \ru SPAN          \cite{Hu2020span}          & editing      & \checkmark & HP filters           & \checkmark & ~          \\
            \ru CR-CNN        \cite{Yang2020constrained} & editing      &            & trainable HP filter  & \checkmark &            \\
            \ru CAT-Net v2    \cite{kwon2022learning}    & JPEG         & \checkmark & DCT filter           & \checkmark & ~          \\ 
            \ru MVSS-Net      \cite{chen2021image}       & editing      &            & trainable HP filter  & \checkmark & \checkmark \\ 
            \ru IF-OSN        \cite{wu2022robust}        & editing      & \checkmark &    -                 & \checkmark & ~          \\
            \ru PSCC-Net      \cite{liu2022pscc}         & editing      & \checkmark &    -                 & \checkmark & \checkmark \\
            \ru E2E           \cite{Marra2020}           & editing      & \checkmark & noiseprint           & ~          & \checkmark \\
            \bottomrule
            
\end{tabular}
}
\caption{Methods used for comparison. We indicate the artifacts they rely on and the input type, if they work only on RGB and/or on noise features extracted through high-pass (HP) filtering. In addition, we also indicate if they have been designed for localization (L), detection (D) or both.}
\label{tab:summary}
\end{table}

\subsection{State-of-the-art comparison}
To ensure a fair comparison we considered only methods with code and/or pre-trained models publicly available on-line 
and run them on the selected testing datasets.
Moreover, to avoid biases, we included only the approaches trained on datasets disjoint from the test datasets.
Eventually, we included two model-based methods:
ADQ \cite{bianchi2011improved} that relies on JPEG artifacts, 
Splicebuster \cite{Cozzolino2015splice} that exploits noise artifacts;
and 11 deep learning-based methods:
EXIF SelfConsistency \cite{Huh2018},
Constrained R-CNN \cite{Yang2020constrained},
RRU-Net \cite{Bi2019rru},
ManTraNet \cite{Wu2019mantra},
SPAN \cite{Hu2020span},
AdaCFA \cite{Bammey2020adaptive},
E2E \cite{Marra2020},
CAT-Net v2 \cite{kwon2022learning},
IF-OSN \cite{wu2022robust},
MVSS \cite{chen2021image},
PSCC-Net \cite{liu2022pscc}, 
Noiseprint \cite{Cozzolino2020noise}.
A brief summary of these methods is provided in Tab.~\ref{tab:summary}.

\vspace{2mm}
\noindent
\bb{Localization results.}
In Tab.~\ref{tab:comparisonF1} we show the pixel-level localization performance.
Our method provides the best F1 performance, on average, and is the best or second best on all datasets,
which testifies of a remarkable generalization ability across manipulations.
In fact, it performs well also on OpenForensics (GAN-based local manipulations), where most other methods fail catastrophically, except CAT-Net v2, as well as CocoGlide (diffusion-based local manipulations).
Thanks to the use of \NAMERES{}, with its digital history-based training, our method keeps working well on all the datasets.

\vspace{2mm}
\noindent
\bb{Detection results.}
Detection results are shown in Tab.~\ref{tab:comparisonAUC}. 
Note that we also consider methods that were not explicitly designed for this task, in which case we use the maximum of the localization map as the detection statistic, as it works better than the mean value. 
\NAME{}  is the best performer on most datasets, and has the best average performance both in terms of AUC and Accuracy.
On the contrary, many methods exhibit a very poor performance, close to random guessing (0.5).
This phenomenon is especially acute for accuracy, which is highly sensitive to the choice of threshold (see supplementary).
Indeed, lacking a suitable calibration dataset, setting the right threshold is a difficult problem, as also shown in \cite{dong2022mvss}. 
Unlike most competitors, our approach guarantees an accuracy of almost 80\% even in this challenging case.

\vspace{2mm}
\noindent
\bb{Robustness analysis.}
In this section we carry out a robustness analysis on images impaired by compression and resizing.
To this end, we use three datasets uploaded on Facebook and Whatsapp - two provided in \cite{wu2022robust} and our CocoGlide.
For compactness, in Tab.~\ref{tab:robustnessF1_fixTH} we compare results only with the top three competitors according to the F1 performance (fixed threshold) of Tab.~\ref{tab:comparisonF1}: IF-OSN, CAT-Net v2 and MVSS-Net.
\NAME{} performs consistently better than all competitors, 
even though IF-OSN was specifically proposed to deal with images transmitted via social networks,
while the gap with respect to CAT-Net v2 and MVSS-Net widens significantly.

\setlength{\tabcolsep}{2pt}
\begin{table}
	\centering
	\small
	\scalebox{\scalefactortab}{
	\begin{tabular}{r C{0.8cm}C{0.8cm}C{0.8cm} C{0.8cm}C{0.8cm}C{0.8cm}}
		\toprule
		\ru     \multirow{2}{*}{\bb{Method}}    & \mcb{2}{CASIAv1}     & \mcb{2}{DSO-1}         & \mcb{2}{CocoGlide}     \\ 
		\ru                                     & Fb       & Wa        & Fb        & Wa         & Fb        & Wa         \\
		\cmidrule(lr){1-1}                       \cmidrule(lr){2-3}      \cmidrule(lr){4-5}       \cmidrule(lr){6-7} 
		\ru  IF-OSN     \cite{wu2022robust}     & .513     & .524      & .484      & .395       & .406      & .404       \\
		\ru  CAT-Net v2 \cite{kwon2022learning} & .681     & .508      & .310      & .247       & .447      & .443       \\
		\ru  MVSS-Net   \cite{chen2021image}    & .469     & .444      & .356      & .308       & .347      & .351       \\ 
		\ru  \NAME~ (ours)                      & \bb{.716}& \bb{.713} & \bb{.685} & \bb{.465}  & \bb{.460} & \bb{.461}  \\  
		\bottomrule       
	\end{tabular}
	}
	\caption{Pixel-level F1 performance (using fixed threshold) on datasets uploaded on Facebook (Fb) and WhatsApp (Wa).
 } 
	\label{tab:robustnessF1_fixTH}
\end{table}

\begin{figure}[b]
	\centering
    \includegraphics[page=1, width=0.95\linewidth]{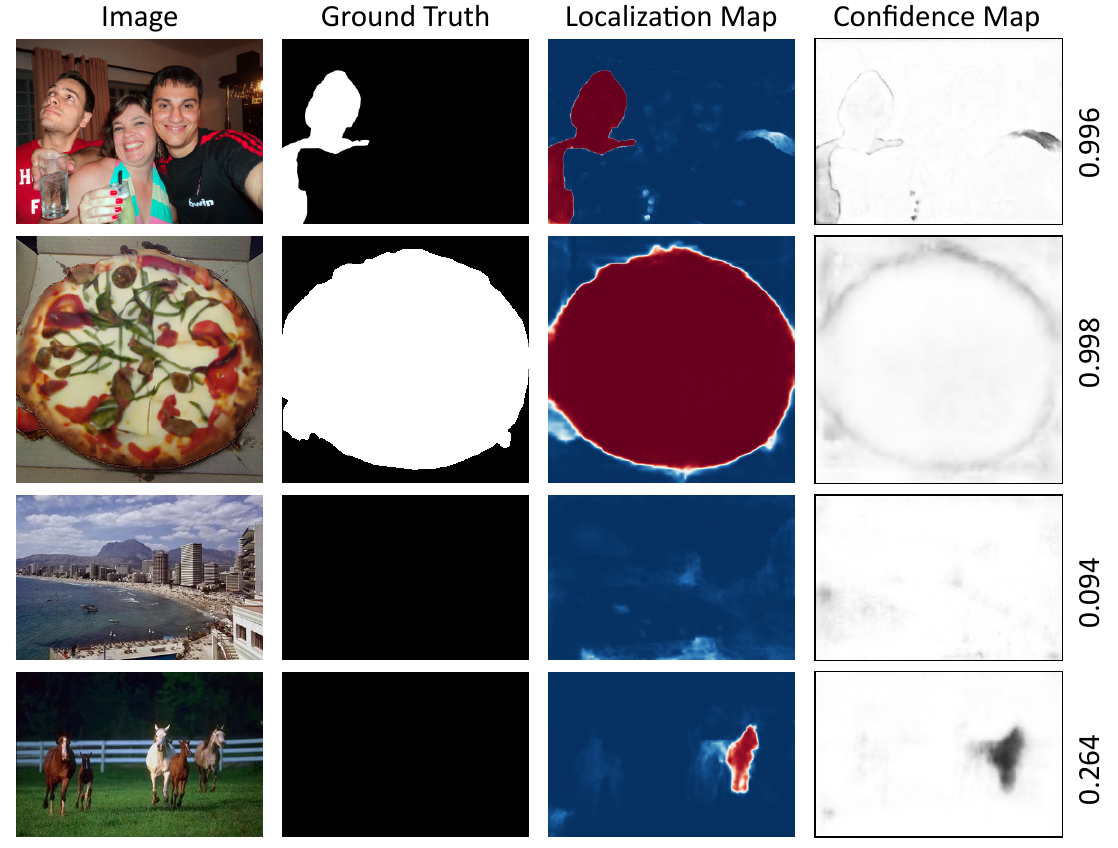}
	\caption{Fake images (top) and pristine images (bottom). In the last row, we show the confidence map that is able to correct the error in the localization map, hence improving the global integrity score (shown on the right).}
	\label{fig:examples}
\end{figure}

\vspace{2mm}
\noindent
\bb{Qualitative comparisons.}
In Fig.~\ref{fig:examples} we also show some visual results in order 
to gain a better insight into the quality of the image localization maps and corresponding confidence maps.
Together with some fakes,
we show some real images, for which the localization map can be erroneous.
In these cases we show the anomaly map, which often presents some hot spots that could lead to false positives. 
Such errors are avoided in detection thanks to the additional confidence map. The user may inspect all these pieces of information to carry out further analyses. More qualitative results are shown in Supplementary.

\begin{table}
    \centering
    \small
    \scalebox{\scalefactortab}{
    \setlength{\tabcolsep}{3pt} 
	\begin{tabular}{lC{0.7cm}C{0.7cm}C{0.7cm}C{0.7cm}C{0.7cm}C{0.7cm}} 
	\toprule
	    \ru                     & \mcb{2}{Original}     & \mcb{2}{Res}          & \mcb{2}{Res\&Cmp}     \\ \cmidrule(lr){2-3} \cmidrule(lr){4-5} \cmidrule(lr){6-7}
	    \ru  \bb{Version}       &      F1   &      AUC  &      F1   &      AUC  &      F1   &      AUC  \\ \cmidrule(lr){1-1} \cmidrule(lr){2-3} \cmidrule(lr){4-5} \cmidrule(lr){6-7}
	    \ru  Noiseprint         &     .706  &     .547  &     .375  &     .483  &     .342  &     .468  \\ 
	    \ru  \NAMERES{}         &     .877  &     .913  &     .666  &     .745  &     .435  &     .566  \\ 
	    \ru  SegFormer (NP++)   &     .974  &     .967  &     .925  & \bb{.966} &     .649  &     .703  \\ 
	    \ru  SegFormer (RGB)    &     .917  &     .903  &     .780  &     .792  &     .756  & \bb{.786} \\ 
	    \ru  \NAME{} (NP++, RGB)& \bb{.982} & \bb{.974} & \bb{.937} &     .944  & \bb{.765} &     .730  \\
		\bottomrule
	\end{tabular}
	}
	\caption{Ablation results. Pixel-level F1 performance (using best threshold) and image-level AUC on original images, resized (Res) and resized and recompressed (Res\&Cmp).}
	\label{tab:ablation1}
\end{table}

\subsection{Ablation study}
In order to assess the individual impact of all design choices of our approach,
we consider a simple baseline, the noiseprint-based method proposed in \cite{Cozzolino2020noise}, and add the new key components one at a time.
Experiments are carried out on a dataset of 1000 manipulated images built by 
downloading pristine images from the web and editing them locally, so as to simulate a realistic scenario.
Tab.~\ref{tab:ablation1} shows the results (F1 and AUC) for 
the noiseprint baseline,
the version with \NAMERES{}, 
the method which includes transformer-based segmentation 
(using as inputs only RGB and only NP++), 
and the proposed method with joint analysis of \NAMERES{} and RGB image. 
We also perform this analysis after resizing all images and after resizing and compressing them.
The case with strongly impaired images is more challenging, but the proposed method keeps providing a good performance.
In general, the inclusion of high-level segmentation information seems to provide the largest improvement,
justifying our focus on all-round clues.

With Tab.~\ref{tab:ablation2}
we study the effect of using the cross-entropy loss alone or jointly with the dice loss, with and without online augmentation with compressed and resized images.
On the original data, results (F1 with best threshold) remain pretty stable in all cases.
With resized and compressed data, instead, the joint use of cross-entropy and dice loss proves important, especially together with augmentation.

Finally, in Tab.~\ref{tab:ablation3} we consider image-level detection 
and compare our method with two simplified versions that rely on a single global feature, the mean or the maximum of the anomaly map.
First of all, it is clear that the mean is a poor decision statistic, 
and much better AUC results can be obtained by just switching to the maximum.
However, even the maximum turns out to be almost useless without a calibration process that helps select a good decision threshold.
So, in terms of accuracy, the feature vector used in the proposed method provides a large competitive advantage.

\begin{table}
    \centering
    \small
    \scalebox{\scalefactortab}{
    \setlength{\tabcolsep}{3pt}
	\begin{tabular}{lC{0.7cm}C{0.7cm}C{0.7cm}C{0.7cm}C{0.7cm}C{0.7cm}C{0.7cm}}
	\toprule
        \ru          &           & \mcb{2}{Original}      &\mcb{2}{Res}             &\mcb{2}{Res\&Cmp}    \\ 
        \ru \bb{Loss}& \bb{Aug}  & best       & fixed     & best        & fixed     & best      & fixed   \\     \cmidrule(lr){1-2} \cmidrule(lr){3-4} \cmidrule(lr){5-6} \cmidrule(lr){7-8}
		\ru  CE      &           & .980       & .926      & .912        & .824      & .583      & .397    \\
		\ru  CE      & \checkmark& .981       & .929      & .885        & .781      & .575      & .575    \\
		\ru  CE+DL   &           & .973       & .949      & .865        & .767      & .655      & \bb{.655}  \\
		\ru  CE+DL   & \checkmark& \bb{.982}  & \bb{.970} & \bb{.937}   & \bb{.902} & \bb{.765} & .627       \\ 
		\bottomrule
	\end{tabular}
	}
	\caption{Localization ablation results: Pixel-level F1 performance (using best and fixed threshold).}
	\label{tab:ablation2}
\end{table}

\begin{table}
    \centering
    \small
    \scalebox{\scalefactortab}{
    \setlength{\tabcolsep}{3pt} 
	\begin{tabular}{C{0.8cm}C{0.8cm}C{0.8cm}C{0.8cm}C{0.8cm}C{0.8cm}C{0.8cm}} 
	\toprule
	    \ru            & \mcb{2}{Original}   & \mcb{2}{Res}        & \mcb{2}{Res\&Cmp}   \\   \cmidrule(lr){2-3} \cmidrule(lr){4-5} \cmidrule(lr){6-7}
        \ru \bb{score} & AUC      & Acc      & AUC      & Acc      & AUC        & Acc    \\   \cmidrule(lr){1-1} \cmidrule(lr){2-3} \cmidrule(lr){4-5} \cmidrule(lr){6-7}
		\ru  mean      & .544     & .510     & .604     & .525     & .592       & .515   \\
		\ru  max       & .974     & .505     & .944     & .515     & .730       & .500   \\
		\ru  Ours      & \bb{.996}& \bb{.905}& \bb{.949}& \bb{.910}& \bb{.740}  & \bb{.675} \\
		\bottomrule
	\end{tabular}
	}
	\caption{Detection ablation results: Image-level AUC and Accuracy (best threshold).}
	\label{tab:ablation3}
\end{table}

\section{Conclusions}
In this paper we introduce \NAME, a novel framework for reliable image forgery detection and localization. It is built upon the extraction of a learned noise-sensitive fingerprint, that enhances the in-camera and out-camera artifacts even in challenging scenarios, such as circulation on social networks. The model also provides a confidence map that represents an indication of possible false alarms on pristine areas. Our extensive experimental results demonstrate that our approach has a good generalizability and is able to localize even unknown manipulations, such as the recent DNN-based ones. Furthermore, it can provide reliable and robust detection results at image level thanks to the introduction of the confidence map.
Our approach has certain limitations. First, it cannot detect fully generated images. Then, we train the anomaly map and detection score in separate phases, requiring full pixel-level supervision. In future work, we would like to explore end-to-end training, allowing partial supervision from only image-level labels. We would also like to evaluate generalization on more recent generative models for local edits~\cite{avrahami2022blendedlatent, gal2022textual}.

\paragraph{Acknowledgment.}
We gratefully acknowledge the support of this research by the Defense Advanced Research Projects Agency (DARPA) under agreement number FA8750-20-2-1004. 
In addition, this work has received funding by the European Union under the Horizon Europe vera.ai project, Grant Agreement number 101070093. It is also supported by a TUM-IAS Hans Fischer Senior Fellowship and by the PREMIER project, funded by the Italian Ministry of Education, University, and Research within the PRIN 2017 program.
Finally, we would like to thank Chris Bregler for useful discussions and support.

\begin{appendix}
\section*{Supplementary Document}
\input{appendix}
\end{appendix}

\vfill\pagebreak

{\small
\bibliographystyle{ieee_fullname}
\bibliography{egbib}
}

\end{document}

%% file: appendix.tex
In this appendix, we report the details of our approach (Sec. \ref{Imp})
and of the datasets used in the experiments (Sec. \ref{datasets}).
Then, we include additional results to prove the robustness capability of our method (Sec. \ref{robustness})
and its ability to provide good results also for the detection task (Sec. \ref{detection}).
Furthermore, we show qualitative results by means of localization and confidence maps and finally we present failure cases (Sec. \ref{qualitative}). Code is publicly available at \url{https://grip-unina.github.io/TruFor/}.

\section{Implementation details}
\label{Imp}

\begin{figure*}
    \centering
    \includegraphics[page=4, width=1.0\linewidth, trim=0 300 0 0,clip]{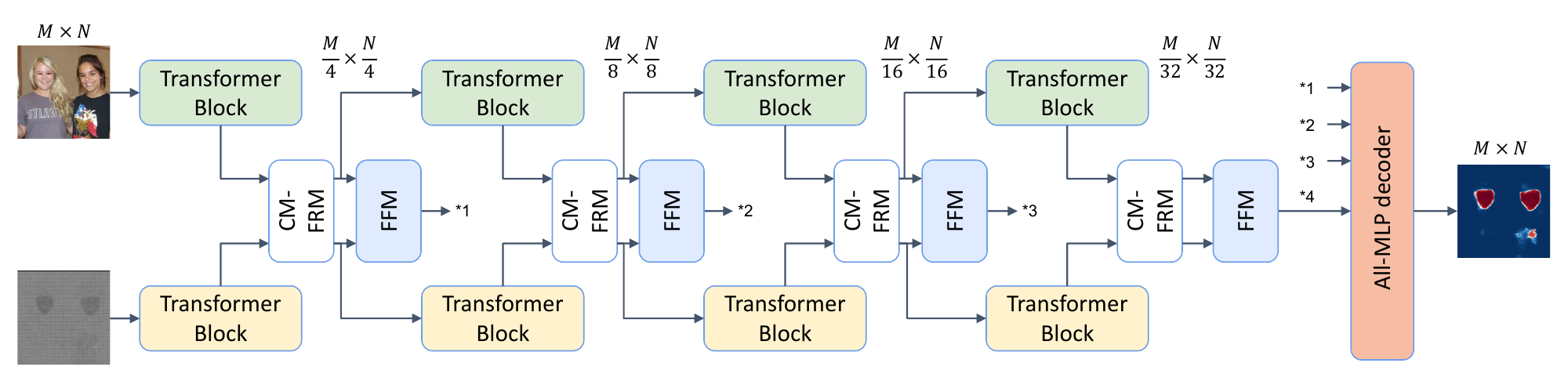}
    \captionof{figure}{Anomaly localization network.}
    \label{fig:arc}
\end{figure*}

\noindent
\bb{Architecture.}  
The anomaly localization network is
shown in Fig.~\ref{fig:arc}. The feature extraction backbone in the encoder is based on a transformer-based segmentation architecture~\cite{xie2021segformer}. The RGB and the \NAMERES{} feature maps are combined using a Cross-Modal Feature Rectification Module (CM-FRM)~\cite{liu2022cmx}.
Each feature extraction branch has 4 Transformer blocks, and a CM-FRM block between each transformer block.
The Transformer blocks are based on the Mix Transformer encoder B2 (MiT-B2) proposed for semantic segmentation and are pretrained on ImageNet, as suggested in \cite{xie2021segformer}.
The Mix Transformer encoder includes self-attention mechanisms and channel-wise operations. It relies on spatial convolutions and not on positional encodings.
This is important in order to work with images of any size and to obtain a localization map with the same resolution as the input image.  

The CM-FRM block exploits the interactions between the image semantic (RGB) and residual (\NAMERES{}) features.
It performs channel-wise and spatial-wise rectifications, which 
consists of a weighted sum of the feature map of both branches. 
The weights are calculated along the channel dimension and the spatial dimension separately, combining both feature maps.
The Feature Fusion Module (FFM) uses an efficient cross-attention mechanism, without positional encoding, to merge the feature maps of \NAMERES{} and RGB image
and the outputs of the four FFMs represent the input of the decoder.
We use the All-MLP decoder proposed in \cite{xie2021segformer}, which is a lightweight architecture formed by only 1$\times$1 convolution layers and bi-linear up-samplers. The decoder for the confidence map has the same All-MLP architecture.
The forgery detector network takes as input the  pooled features from anomaly and confidence maps, and consists of 2 fully connected layers with RELU activation: $8D \rightarrow 128D \rightarrow 1D$ output.

Experiments have been conducted using one NVIDIA RTX A6000 GPU.
Training times for each phase are 6.5 days, 6 days, 2 days, respectively.
The inference time is about 1.17 sec for an image of 3.2 megapixels.
As for the model size, the number of parameters for TruFor is 68.7M, that are less than those used by the top three competitors:
CAT-Net v2 (114.3M), MVSS-Net (146.9M) and IF-OSN (128.8M).

\begin{figure}
	\centering
    \includegraphics[page=3, width=1.0\linewidth, trim=145 135 145 0,clip]{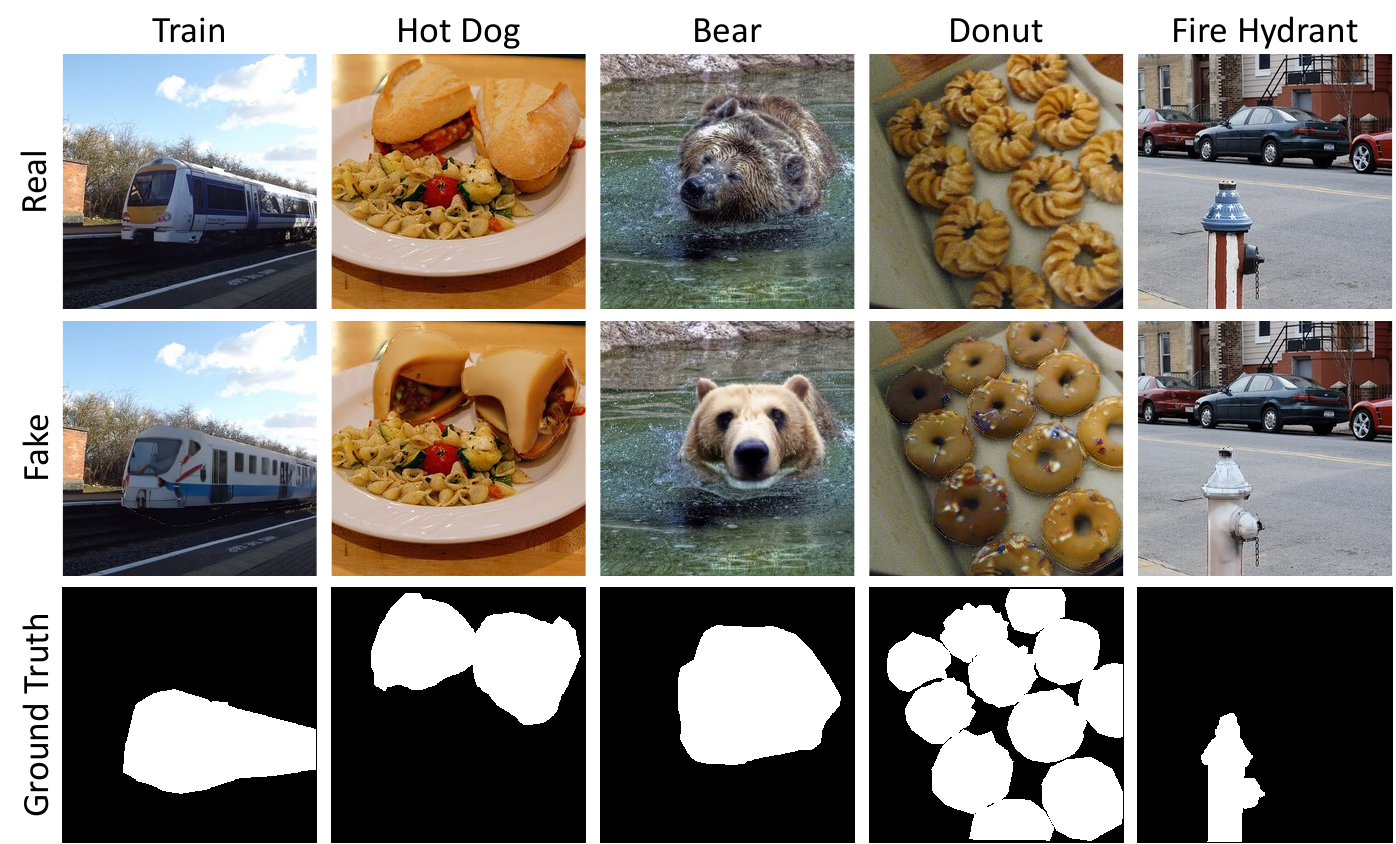}
	\caption{Some examples of real and manipulated images and related reference maps from the CocoGlide dataset. For each image we indicate the prompt that drives the synthetic generation.
 }
	\label{fig:glide_examples}
\end{figure}

\vspace{2mm}
\noindent
\bb{\NAMERES{} training.}
For \NAMERES{} training, each batch includes 160 patches of $64\times 64$ pixels. These patches are obtained from 5 camera models and 4 different images for each camera model. The resulting 20 images are subjected to 4 different editing histories, which are a combination of random resizing, compression and contrast/brightness adjustments, 
for a total of 512 possible editing histories.
Training is performed for a total of 50 epochs, and each epoch includes 8800 training steps. An Adam optimizer is used with an initial learning rate of 0.001, that is reduced by 10 times every 10 epochs.

\vspace{2mm}
\noindent
\bb{Localization and detection training.}
For localization and detection tasks we adopted the datasets used for training and validation also used in \cite{kwon2022learning}, which
comprises both pristine and fake images with the corresponding reference maps. 
The input image is cropped to $512\times 512$ during training. Details of the dataset are reported in Tab.~\ref{tab:db}. To avoid biases due to an imbalance in training dataset size, we sample each dataset equally for each training epoch.
The networks are trained for 100 epochs with a batch size equal to 18 and a learning rate that starts with 0.005 and decays to zero. An SGD optimizer is used with a momentum of 0.9.
Before \NAMERES{} extraction, 
we apply the following augmentations on RGB inputs:
resizing in the range [0.5 - 1.5] and JPEG compression with quality factor from 30 to 100. 

\section{Datasets}
\label{datasets}

To ensure that Noiseprint++ is trained on unaltered images, we verified that for each camera model, all collected images have the same resolution, are in JPEG format with the same quantization matrix and that no photo editing software is present in the metadata (e.g. photoshop, gimp).

As for the anomaly localization and detection,
the datasets used for training and testing are reported in Tab.~\ref{tab:db}. Training includes CASIA v2 \cite{Dong2013}, FantasticReality \cite{Kniaz2019}, IMD2020 \cite{Novozamsky2020} and a dataset of manipulated images created by \cite{kwon2022learning} by applying splicing and copy-move using either COCO \cite{lin2014microsoft} training set or RAISE \cite{DangNguyen2015raise} as a source and object masks from COCO as target regions. 
For OpenForensics \cite{Le2021OpenForensics} and NIST16 \cite{Guan2019MFC}, we evaluate the performance on a test subset of 2000 images (out of 19,000) and 160 images, respectively. The latter choice follows the common train/test split that most of the recent works apply \cite{Hu2020span, Yang2020constrained, zhou2018learning, wang2022objectformer}.
CocoGlide is a manipulated dataset generated by us using the COCO validation dataset \cite{lin2014microsoft}. We extract  $256 \times 256$ pixel crops and then use an object mask and its corresponding label as the forgery region and the text prompt that are fed to GLIDE \cite{nichol2021glide}. In this way, we generated new synthetic objects of the same category for a total of 512 manipulated images. Some examples are shown in Fig.~\ref{fig:glide_examples}. Note that we avoided overlap with \cite{kwon2022learning}, since CocoGlide is based on images from the validation set, while the tampered COCO dataset from the training set.

\setlength{\tabcolsep}{3pt}
\begin{table}[b]
\centering
{\footnotesize
\begin{tabular}{crcccc}
  \toprule
  \ru                                          &              & \mcb{2}{Number of images}               & \mcb{2}{Manipulation}              \\
  \cmidrule(lr){3-4} \cmidrule(lr){5-6}
  \ru &  Name [ref]~~~~                         & Real        & Fake                & Sp          & CM   \\
  \midrule \ru 
      & CASIA v2         \cite{Dong2013}~~~~             &  7491 & 5105 & \checkmark & \checkmark \\
  \ru & FantasticReality \cite{Kniaz2019}~~~~            & 16592 & 19423 & \checkmark & ~         \\
  \ru & IMD2020          \cite{Novozamsky2020}~~~~       &   414 & 2010 & \checkmark & \checkmark \\
  \ru & tampered COCO    \cite{kwon2022learning}~~~~     &     - & 400K & \checkmark & \checkmark \\
  \ru & tampered RAISE   \cite{kwon2022learning}~~~~     & 24462 & 400K &  ~  &  \checkmark  \\
  \midrule \ru 
      & CASIA v1+        \cite{dong2022mvss}~~~~         &   800 & 921  & \checkmark & \checkmark \\
  \ru & Coverage         \cite{wen2016coverage}~~~~      &   100 & 100  & ~          & \checkmark \\
  \ru & Columbia         \cite{hsu2006detecting}~~~~     &   183 & 180  & \checkmark & ~          \\ 
  \ru & NIST16           \cite{Guan2019MFC}~~~~          &   160 & 160  & \checkmark & \checkmark \\ 
  \ru & DSO-1            \cite{Carvalho2013}~~~~         &   100 & 100  & \checkmark & ~          \\
  \ru & VIPP             \cite{bianchi2012image}~~~~     &    69 & 69   & \checkmark & \checkmark \\
  \ru & OpenForensics    \cite{Le2021OpenForensics}~~~~  &     - & 2000 & \checkmark & ~          \\
  \ru & CocoGlide        ~~~~                            &   512 & 512  & \checkmark & ~          \\
            
  \bottomrule
\end{tabular}
}
\caption{List of datasets used for training and testing (Sp=splicing, CM=copymove).}
\label{tab:db}
\end{table}

\begin{figure*}
    \centering
    \includegraphics[page=1,width=0.245\linewidth]{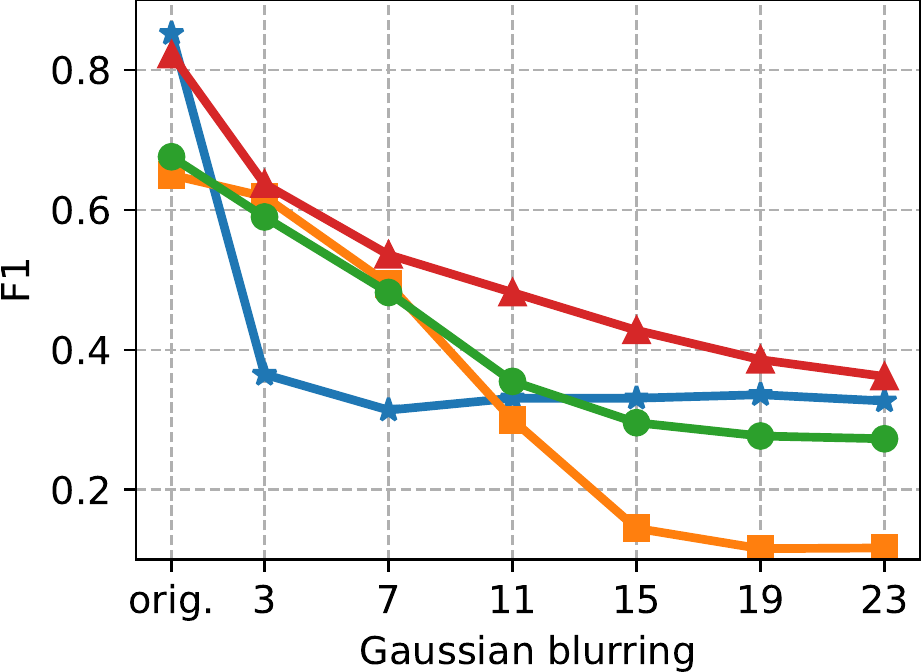} 
    \includegraphics[page=2,width=0.245\linewidth]{images/pixel_level_ablation.pdf} 
    \includegraphics[page=3,width=0.245\linewidth]{images/pixel_level_ablation.pdf}
    \includegraphics[page=4,width=0.245\linewidth]{images/pixel_level_ablation.pdf}
    \includegraphics[page=5,width=0.55\linewidth, trim=40 93 40 80]{images/pixel_level_ablation.pdf}
    \caption{Robustness analysis against different processing operations on CASIA v1. Pixel-level F1 performance (best threshold) is shown.}
    \label{fig:robustness}
\end{figure*}
\begin{table*}
	\centering
	\small
	\scalebox{0.90}{
	\setlength{\tabcolsep}{2pt}
	\begin{tabular}{rC{0.7cm}C{0.7cm}C{0.7cm}C{0.7cm}C{0.7cm}C{0.7cm}C{0.7cm}C{0.7cm}C{0.7cm}C{0.7cm}C{0.7cm}C{0.7cm}C{0.7cm}C{0.7cm}C{0.7cm}C{0.7cm}C{0.7cm}C{0.7cm}C{0.7cm}C{0.7cm}}
\toprule
\ru                                     & \mcb{4}{CASIA v1}                      & \mcb{4}{Columbia}                   & \mcb{4}{DSO-1}                        & \mcb{4}{NIST16}    & \mcb{4}{AVG}    \\ 
 \mcb{1}{\ru\bb{Method}}                
    & Fb      & Wa      & Wb      & Wc      & Fb      & Wa      & Wb     & Wc     & Fb      & Wa      & Wb      & Wc      & Fb      & Wa      & Wb      & Wc      & Fb      & Wa      & Wb      & Wc       \\ 
\cmidrule(l){1-1} \cmidrule(lr){2-5} \cmidrule(lr){6-9} \cmidrule(lr){10-13}  \cmidrule(lr){14-17} \cmidrule(lr){18-21}   
\ru  IF-OSN     \cite{wu2022robust}     
    & .513    & .524    & .507    & .454    & .741	  & .752	&  .756   &  .760   & .484    & .395    & .416    & .414    &  .315	  &  .302	&  .292	  & .282   & .513 & .493 & .493 & .478  \\
\ru  CAT-Net v2 \cite{kwon2022learning}
    & .681    & .508    & .469    & .206    &\bb{.964}&\bb{.952}&\bb{.958}&\bb{.903}& .310    & .247    & .240    & .237    &  .219	  &  .238	&  .243	  & .244   & .544 & .486 & .478 & .398  \\
\ru  MVSS-Net   \cite{chen2021image}
    & .469    & .444    & .480    & .339    & .752    & .747	&  .758   &  .752   & .356    & .308    & .354    & .329    &  .305   &  .252   &  .300   & .269   & .471 & .438 & .473 & .422  \\ 
\ru  \NAME~ (ours)
    &\bb{.716}&\bb{.713}&\bb{.676}&\bb{.615}& .797    & .798    &  .835   &  .820   &\bb{.685}&\bb{.465}&\bb{.515}&\bb{.469}&\bb{.338}&\bb{.384}&\bb{.308}&\bb{.358}&\bb{.634}&\bb{.590}&\bb{.584}&\bb{.566}\\  
\bottomrule       
	\end{tabular}

	}
	\caption{Pixel-level F1 performance (fixed threshold) on datasets uploaded on Facebook (Fb), WhatsApp (Wa), Weibo (Wb), WeChat (Wc).
 } 
	\label{tab:socialmedia}
\end{table*}

\section{Additional robustness analysis}
\label{robustness}

In this Section we include additional experiments to show the ability of our method to be robust to different forms of degradations and compare them with those obtained by the top performers \cite{wu2022robust, kwon2022learning, chen2021image}.
We apply the following transformations on the CASIA v1 dataset: gaussian blur (varying the kernel size), gaussian noise (varying the standard deviation), gamma correction (varying the power factor) and JPEG compression (varying the quality level).
The results are shown in Fig.~\ref{fig:robustness}.
We can observe that our method is more robust than the state-of-the-art irrespective of the type of degradation.

We also check robustness to other social media networks, beyond those already considered in the main paper, i.e. Facebook and Whatsapp (Tab.~4 in main paper). More specifically, we use the whole dataset proposed in \cite{wu2022robust}, 
where images from some standard forensic datasets, CASIA v1\footnote{Actually, we used the v1+ version \cite{dong2022mvss}, 
where real images of v1 (shared with v2 and present in our training set) are replaced by images from the COREL dataset \cite{Wang2021simplicity}.} \cite{Dong2013}, Columbia \cite{hsu2006detecting}, DSO-1 \cite{Carvalho2013} and NIST16 \cite{Guan2019MFC}, were also uploaded on Weibo and WeChat. Results are presented in Tab.~\ref{tab:socialmedia} and show a consistent gain over all the different datasets and social platforms except on Columbia, where CAT-Net v2 achieves better performance. On average however, we have a gain of around 16\%, 19\%, 18\% and 18\% with respect to the second best on Facebook, Whatsapp, Weibo and WeChat, respectively.

\begin{table}
    \centering
    \small
\scalebox{0.90}{
\setlength{\tabcolsep}{0pt} 
    \begin{tabular}{lC{1.45cm}C{1.45cm}C{1.45cm}C{1.45cm}C{1.45cm}} 
    \toprule
\ru Method       & Columbia & Coverage & CASIA v1 & NIST16 & Avg  \\ \midrule
\ru ManTraNet    & .824    & .819    & .817    & .795 & .814 \\
\ru SPAN         & .936    & .922    & .797    & .840 & .874 \\
\ru PSCCNet      &\bb{.982}& .847    & .829    & .855 & .878\\
\ru ObjectFormer & .955    &\bb{.928}& .843    & .872 & .900 \\
\ru \NAME~       & .947    & .925    &\bb{.957}&\bb{.877}&\bb{.927}\\
\bottomrule
\end{tabular}
}
    \caption{Pixel-level AUC, for the comparisons the values are taken from Tab.~1 of \cite{wang2022objectformer}}
    \label{tab:objectformer}
\end{table}

\bb{Comparison with ObjectFormer}. Note that an exhaustive and equitable comparison with \cite{wang2022objectformer} is not feasible as they do not provide their trained model. We provide a pixel-level comparison of localization performance in Tab.~\ref{tab:objectformer} using values for~\cite{wang2022objectformer, liu2022pscc} from the paper. Our method is competitive or better than~\cite{wang2022objectformer} across various test datasets, and outperforms that on average.

\section{Additional detection results}
\label{detection}

In this Section, we give some more insights on the image level detection performance of our method. 
We first investigate the role of the confidence map in the detection strategy. In Tab.~\ref{tab:ablation4}, we perform an ablation where we observe substantial improvements with the confidence maps both in terms of AUC and Accuracy.

\begin{figure}
    \centering
    \includegraphics[width=0.95\linewidth]{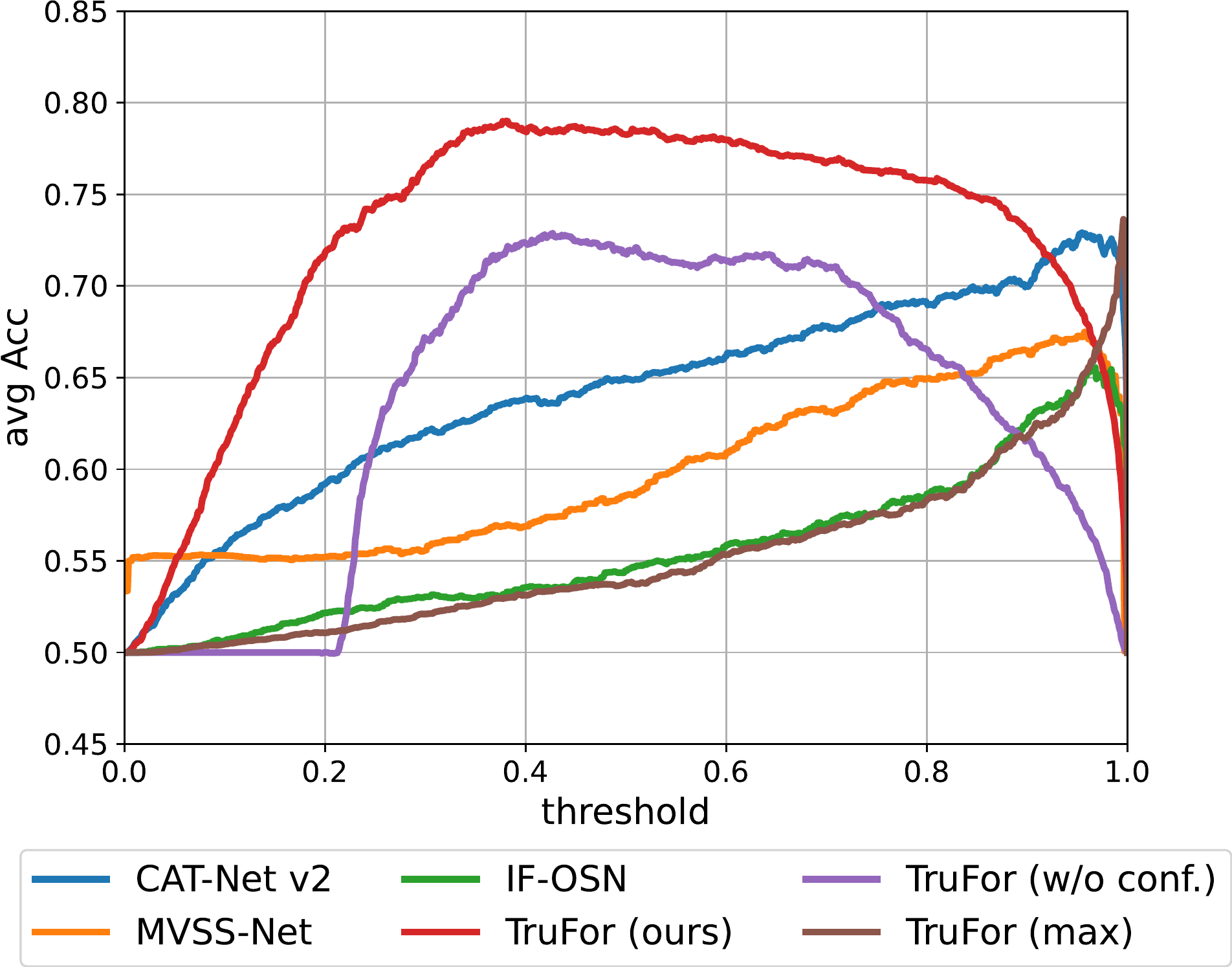}
    \caption{Image-level Detection Accuracy as a function of detection score threshold (averaged over 7 test datasets).}
    \label{fig:decres}
\end{figure}

Image-level metrics require calibrating the detection score for a particular dataset (or certain methods fine-tune on specific datasets~\cite{wang2022objectformer}).  
In Tab.~$2$ of the main paper, we report the balanced accuracies evaluated on seven datasets, and the average of them considering a fixed threshold equal to 0.5. For methods that do not provide an explicit detection score, we use max pooling on the localization map. 
In Fig.~\ref{fig:decres}, we show the accuracy (averaged over the seven datasets) as a function of the threshold. One can observe the accuracy of other methods, which rely on max pooling, increase with higher thresholds - this is indicative of many false positives in the localization maps from these methods. In contrast, our method combines various confidence weighted pooling statistics, making it more robust.  

Table \ref{tab:decres} shows the true negative rate (TNR), true positive rate (TPR), and average accuracy considering both a fixed threshold of 0.5 and the best threshold for each technique. We can notice that using a fixed threshold with our method we can significantly decrease the false alarms rate (around 80\% lower) at the cost of increasing miss detection (around 30\% higher), by achieving an average improvement in terms of accuracy of 25\%. All the state-of-the-art approaches have the problem of a high number of false alarms with a best threshold that assumes values almost equal to 1. Also in this experiment where results are averaged on all seven datasets, we can appreciate the importance to include the confidence analysis during detection.

\section{Qualitative results}
\label{qualitative}

\begin{figure}
	\centering
	\includegraphics[page=1, width=1.0\linewidth]{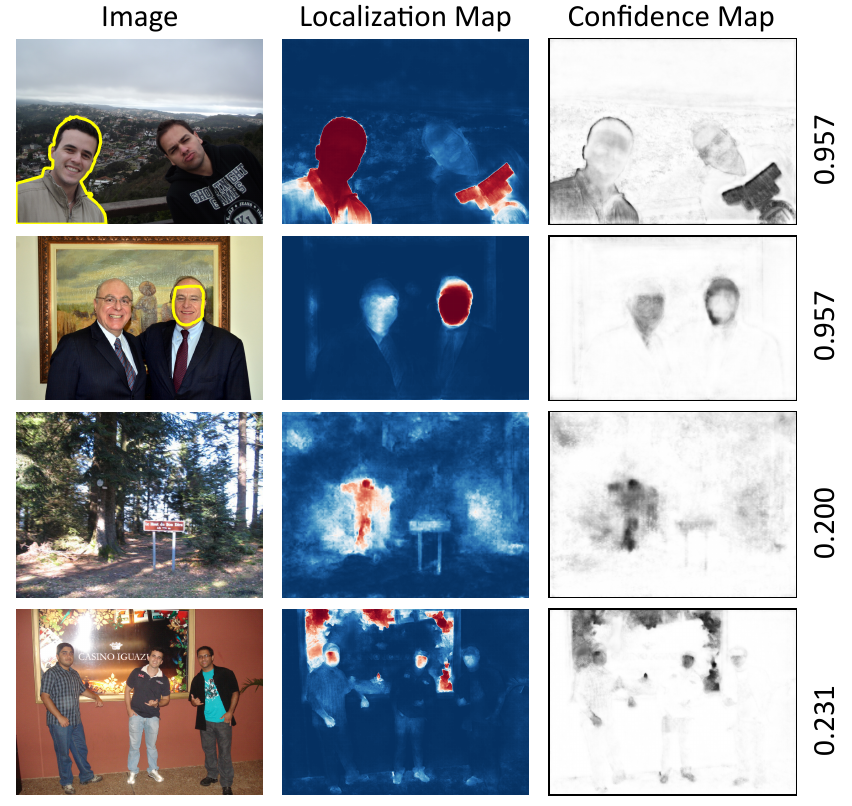}
	\caption{Examples of forged (top) and pristine (bottom) images (forgeries are highlighted in yellow). We show the localization map, the confidence map and the integrity score. For real images despite the anomaly map presenting some false alarms, the confidence analysis helps to make a correct prediction.
	}
	\label{fig:det_examples}
\end{figure}

\begin{figure}
	\centering
    \includegraphics[page=1, width=1.0\linewidth]{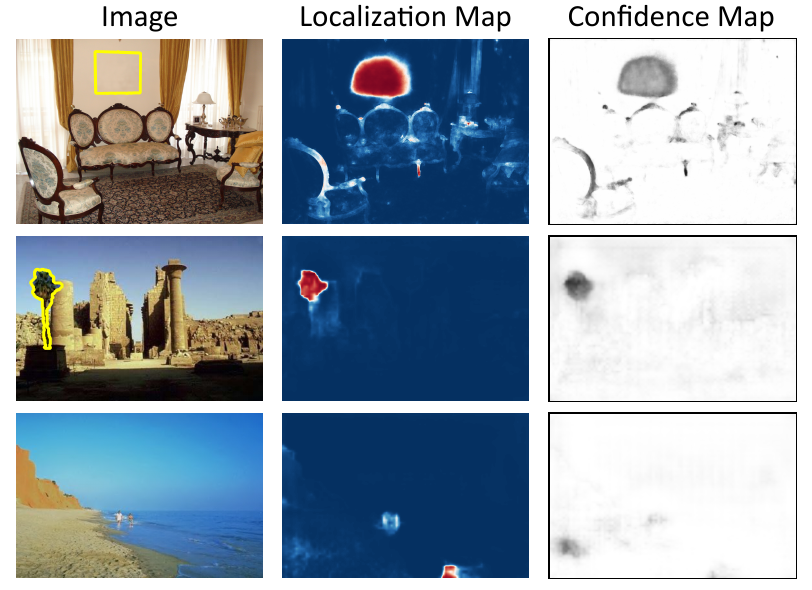}
	\caption{Examples of failure cases on fake and real images (forgeries are highlighted in yellow).}
	\label{fig:err_examples}
\end{figure}

In Fig.~\ref{fig:det_examples} we show some results on fake and pristine images together with the relative confidence map and the final integrity score. We can see that the confidence map can help to correct false positive predictions and provide a more reliable integrity score.
Instead, in Fig.~\ref{fig:err_examples} different failure cases. In the first row, the manipulation was correctly localized, however, the confidence map wrongly hints that it could be a false alarm.
A possible explanation is that the area is very uniform, which can lead to false positives.
A similar situation is presented in the second row, since the plant has a very uniform and dark texture, which misleads the confidence extractor.
Another failure case can be represented by the other way around, where we have a false positive on a pristine image, and the confidence map not correcting it.

\begin{figure*}
	\centering
    \vspace{0.9cm}
    \includegraphics[page=1, width=0.995\linewidth]{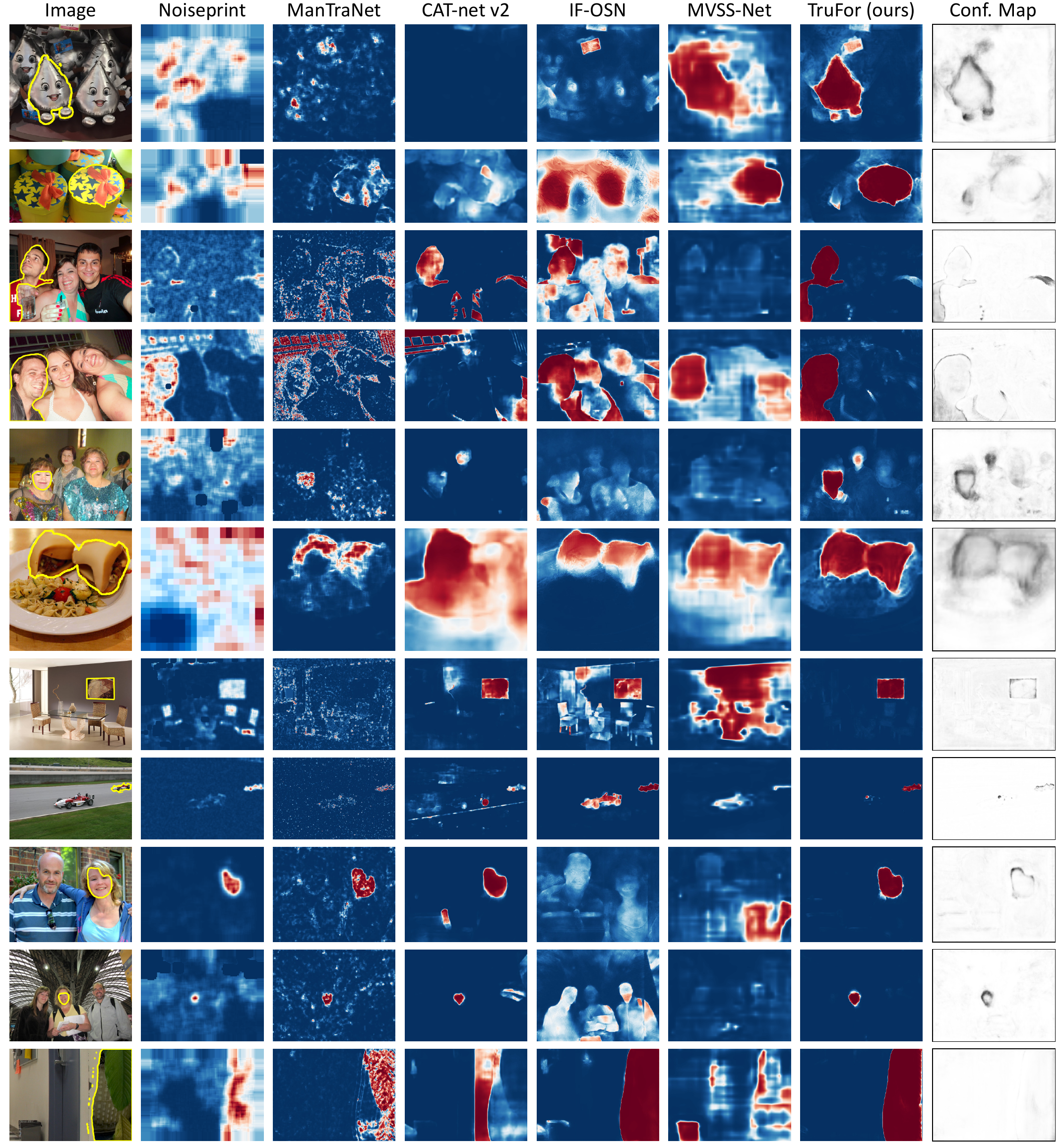}
    \includegraphics[width=0.99\linewidth]{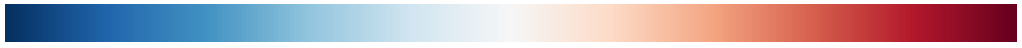}
	\caption{Some qualitative results, compared with the state-of-the-art, on manipulated images (the forged area is outlined in yellow). Dark regions in the confidence map indicate regions of low confidence in the TruFor localization map. }
    \vspace{0.9cm}
	\label{fig:loc_examples}
\end{figure*}

In Fig.~\ref{fig:loc_examples} we show some qualitative results on manipulated images (the forged area is outlined in yellow) and compare with the state-of-the-art.
For these examples, the localized area appears sharper and more accurate than the other methods.
We also add the confidence maps that can tell us the level of reliability of the anomaly maps and remove potential false alarms. Note the dark regions on the boundary of real forgeries - indicating lower confidence in the anomaly label assignment of intermediate regions.

\begin{figure*}
	\centering
    \includegraphics[page=1, width=0.995\linewidth]{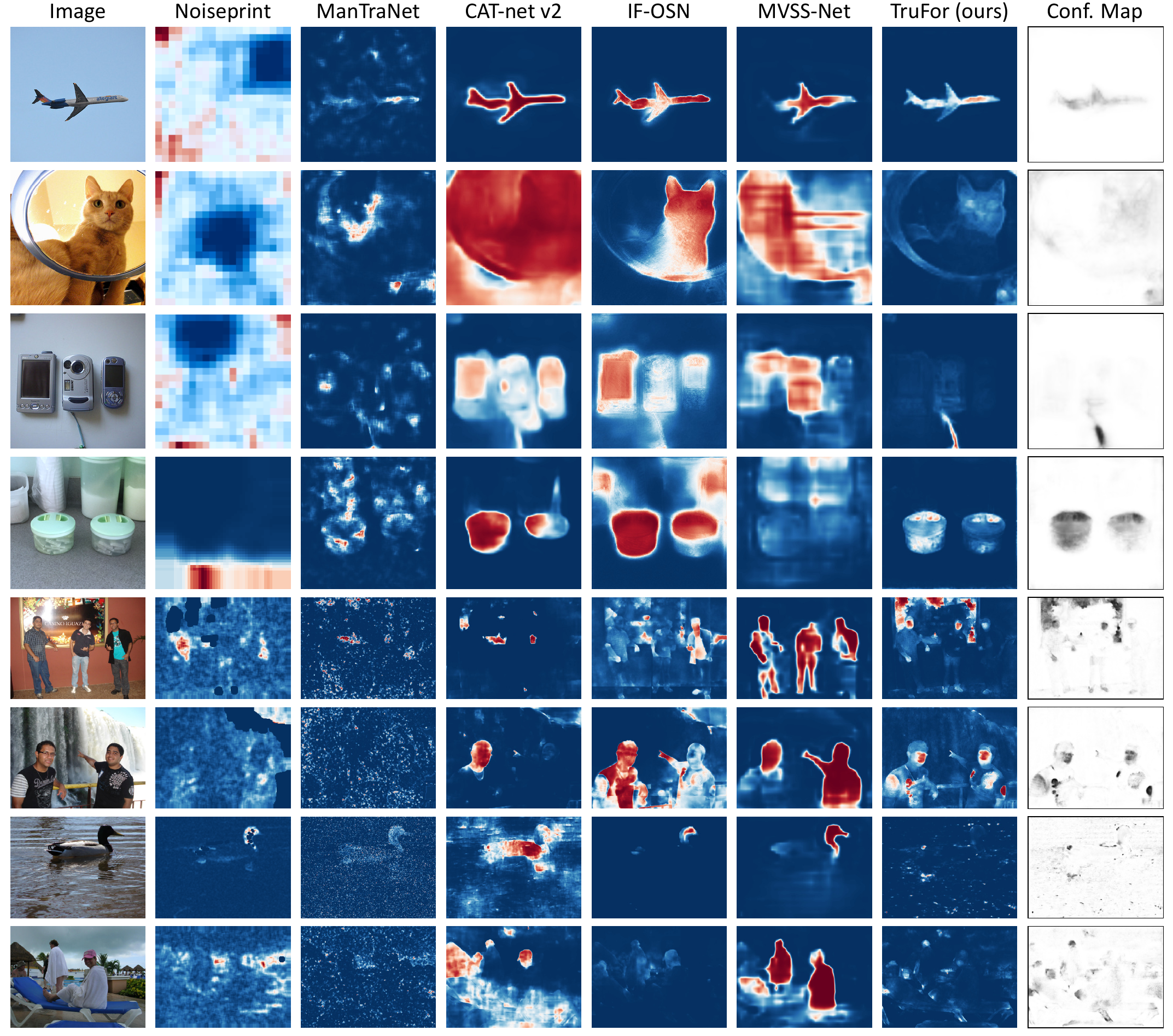}
    \includegraphics[width=0.99\linewidth]{images/cmap.png}
    \caption{Some qualitative results, compared with the state-of-the-art, on pristine images. Dark regions in the confidence map indicate regions of low confidence in the TruFor localization map.}
	\label{fig:exreal}
\end{figure*}

Finally, in Fig.~\ref{fig:exreal}, we show a few examples of false alarms on pristine images. Other methods tend to focus on semantically relevant or highly saturated regions leading to false detections. TruFor's localization maps exhibit a weaker response, and most of these are discarded due to the confidence map, leading to a correct image-level decision.

\begin{table}
    \centering
    \small
    \scalebox{0.90}{
    \setlength{\tabcolsep}{3pt} 
	\begin{tabular}{C{2.2cm}C{0.8cm}C{0.8cm}C{0.8cm}C{0.8cm}C{0.8cm}C{0.8cm}} 
	\toprule
	    \ru                & \mcb{2}{Original}        & \mcb{2}{Res}      & \mcb{2}{Res\&Cmp}   \\   \cmidrule(lr){2-3} \cmidrule(lr){4-5} \cmidrule(lr){6-7}
        \ru                & AUC      & Acc      & AUC      & Acc      & AUC        & Acc   \\  \cmidrule(lr){1-1} \cmidrule(lr){2-3} \cmidrule(lr){4-5} \cmidrule(lr){6-7}
        \ru  w/o conf. map &     .877 &    .785  &     .847 &    .730  &     .719   & .610 \\
		\ru  w conf. map   & \bb{.996}& \bb{.905}& \bb{.949}& \bb{.910}& \bb{.740}  & \bb{.675}      \\
		\bottomrule
	\end{tabular}
	}
	\caption{Ablation image-level results in terms of AUC and accuracy considering the use (or not) of the confidence map.}
	\label{tab:ablation4}
\end{table}

\begin{table}
    \centering
    \small
\scalebox{0.90}{
\setlength{\tabcolsep}{2pt} 
\begin{tabular}{lC{0.8cm}C{0.8cm}C{0.8cm}C{0.8cm}C{0.8cm}C{0.8cm}C{0.8cm}} 
\toprule
\ru              & \mcb{3}{fixed}       & \mcb{4}{best}   \\ \cmidrule(lr){2-4} \cmidrule(lr){5-8}
\ru              &  TNR   & TPR    & Acc     & th     & TNR     & TPR     & Acc  \\
\midrule
\ru CAT-Net v2      &      .416  &       .882  &     .649  &   .955 &     .840  &     .618  &     .729 \\
\ru IF-OSN          &      .182  &       .907  &     .545  &   .968 &     .763  &     .548  &     .656 \\
\ru MVSS-Net        &      .285  &       .886  &     .586  &   .957 &     .806  &     .544  &     .675 \\
\ru \NAME~ (max)    &      .109  &   \bb{.967} &     .538  &   .996 & \bb{.900} &     .573  &     .736 \\
\ru \NAME~ (w/o c.) &      .859  &       .575  &     .717  &   .427 &     .818 &      .640 &      .729 \\
\ru \NAME~    &  \bb{.909} &       .656  & \bb{.783} &   .380 &     .851  & \bb{.729} & \bb{.790} \\
\bottomrule
\end{tabular}
}
	\caption{Detection results: Image-level TNR, TPR, and Accuracy averaged on seven datasets (fixed and best threshold).}
	\label{tab:decres}
\end{table}

